\RequirePackage{fix-cm}
\documentclass[twocolumn]{svjour3}          % twocolumn
\smartqed  % flush right qed marks, e.g. at end of proof
\usepackage{graphicx}
\usepackage{natbib} %Eduardo (Error in bibtex)
\usepackage[strings]{underscore} %Eduardo (Error in DOI)
\usepackage{xcolor} %Eduardo (Text with color)
\usepackage{amsmath} %Eduardo
\usepackage{amssymb} %Eduardo
\usepackage{algorithm} %Eduardo
\usepackage{algpseudocode} %Eduardo
\usepackage{cuted}

\usepackage[colorlinks,
            linkcolor=blue,
            anchorcolor=blue,
            citecolor=blue,
            urlcolor=blue,
            ]{hyperref}
     
\usepackage{etoolbox}

\makeatletter

\patchcmd{\NAT@citex}
  {\@citea\NAT@hyper@{%
     \NAT@nmfmt{\NAT@nm}%
     \hyper@natlinkbreak{\NAT@aysep\NAT@spacechar}{\@citeb\@extra@b@citeb}%
     \NAT@date}}
  {\@citea\NAT@nmfmt{\NAT@nm}%
   \NAT@aysep\NAT@spacechar\NAT@hyper@{\NAT@date}}{}{}

\patchcmd{\NAT@citex}
  {\@citea\NAT@hyper@{%
     \NAT@nmfmt{\NAT@nm}%
     \hyper@natlinkbreak{\NAT@spacechar\NAT@@open\if*#1*\else#1\NAT@spacechar\fi}%
       {\@citeb\@extra@b@citeb}%
     \NAT@date}}
  {\@citea\NAT@nmfmt{\NAT@nm}%
   \NAT@spacechar\NAT@@open\if*#1*\else#1\NAT@spacechar\fi\NAT@hyper@{\NAT@date}}
  {}{}

\makeatother
\newcommand{\textbl}[1]{\color{blue} #1}

\begin{document}

\title{Rule-based Evolving Fuzzy System for Time Series Forecasting: New Perspectives Based on Type-2 Fuzzy Sets Measures Approach}
% \subtitle{Do you have a subtitle?\\ If so, write it here}

\author{Eduardo Santos de Oliveira Marques, Arthur Caio Vargas Pinto, \\ Kaike Sa Teles Rocha Alves, Eduardo Pestana de Aguiar}

\institute{Federal University of Juiz de Fora, Juiz de Fora MG, Brazil}

\date{}

\authorrunning{Eduardo Santos de Oliveira Marques}

\titlerunning{Rule-based Evolving Fuzzy System for Time Series Forecasting: Type-2 Fuzzy Sets Measures Approach}

\maketitle
\begin{sloppypar}
\begin{abstract}
Real-world data contain uncertainty and variations that can be correlated to external variables, known as randomness. An alternative cause of randomness is chaos, which can be an important component of chaotic time series. One of the existing methods to deal with this type of data is the use of the evolving Fuzzy Systems (eFSs), which have been proven to be a powerful class of models for time series forecasting, due to their autonomy to handle the data and highly complex problems in real-world applications. However, due to its working structure, type-2 fuzzy sets can outperform type-1 fuzzy sets for highly uncertain scenarios. We then propose ePL-KRLS-FSM+, an enhanced class of evolving fuzzy modeling approach that combines participatory learning (PL), a kernel recursive least squares method (KRLS), type-2 fuzzy logic and data transformation into fuzzy sets (FSs). This improvement allows to create and measure type-2 fuzzy sets for better handling uncertainties in the data, generating a model that can predict chaotic data with increased accuracy. The model is evaluated using two complex datasets: the chaotic time series Mackey-Glass delay differential equation with different degrees of chaos, and the main stock index of the Taiwan Capitalization Weighted Stock Index - TAIEX. Model performance is compared to related state-of-the-art rule-based eFS models and classical approaches and is analyzed in terms of error metrics, runtime and the number of final rules. Forecasting results show that the proposed model is competitive and performs consistently compared with type-1 models, also outperforming other forecasting methods by showing the lowest error metrics and number of final rules.

\end{abstract}

%% Introduction
\section{Introduction}

During the last years the world has been experiencing an era where datasets are becoming more important to comprehend the current scenario (\citealt{garcia2016big}) and predict the future (\citealt{bontempi2012machine}). Scientists, researchers and organizations are developing algorithms and statistical models to handle these applications, creating computer systems designed to perform specific tasks without being explicitly programmed, which is known as Machine Learning (\citealt{mahesh2020machine}).

Machine Learning (ML) can be defined as a \textit{mechanism for pattern search and building intelligence into a machine to be able to learn, implying that it will be able to do better from its own experience} (\citealt{gollapudi2016practical}). ML models have attracted attention due to their capacity to emulate human intelligence by learning from the neighboring environment (\citealt{el2015machine}). They learn from data, so the greater the number of input data, the more knowledge can be extracted from it and the more effective these models become at problem-solving (\citealt{chen2018lifelong}).

In real-world data, some individual data points are determined partly by perturbations or causal factors that cannot be reduced to any pattern, being called randomness (\citealt{mcallister2003algorithmic}). Randomness is an unobservable property of a generating process, that can only be inferred indirectly from properties of the generator’s output (\citealt{bar1991perception}). Chaos, which can be present even in very simple deterministic systems, is a way to generate randomness (\citealt{farmer1987predicting}). Chaos is a nonlinear deterministic process that makes the data ``look'' random (\citealt{hsieh1991chaos}). An example of data affected by the presence of nonlinearities and chaos is that related to the market, such as financial time series (\citealt{albulescu2021nonlinearities}).

One of the biggest challenges in Machine Learning systems is to have the capability to handle different applications in different environments (\citealt{mitchell1990machine}). The evolving Fuzzy Systems (eFSs) emerge as a powerful class of models that have the capability of learning from data on-the-fly (fast) during online processes in an incremental and mostly single-pass manner (\citealt{lughofer2016evolving}). The eFS is an adaptive process capable of simultaneously learning the model structure and functionality from data flows (\citealt{maciel2017evolving}). This model had its first version with the evolving Takagi-Sugeno (eTS), proposed by \citet{angelov2003line, angelov2004approach}. \citet{lima2010evolving} introduced the evolving Participatory Learning (ePL), a model that combines the eTS structure with the Participatory Learning (PL) concept to control the creation and updating of rules. \citet{maciel2017evolving} created a model that uses the Kernel Recursive Least Square (KRLS) algorithm to update the consequent parameters in the ePL model, the evolving Participatory Learning with Kernel Recursive Least Square (ePL-KRLS). \citet{alves2021novel} introduced a rule-based eFS that introduces the distance correlation in the compatibility measure, named evolving Participatory Learning with Kernel Recursive Least Square and Distance Correlation (ePL-KRLS-DISCO), making the model able to handle data accurately requiring a low number of rules. The compatibility measure is an essential part of the model responsible for creating new rules and eliminating the redundant ones, thus reducing the effect of outliers (\citealt{alves2020enhanced}). This measure acts in two steps: (i) indicating how much each data point is compatible with the current cluster structure and (ii) analyzing when the current cluster structure should be revised upon the arrival of new information (\citealt{maciel2017evolving}).

The notion of a fuzzy set (FS) provides a convenient starting point for the construction of a more general conceptual framework than ordinary sets, having a much wider scope of applicability, particularly in the fields of pattern classification and information processing. Essentially, this framework provides a natural way of dealing with problems where the source of imprecision is the absence of a defined criteria of class membership rather than the presence of random variables (\citealt{zadeh1965fuzzy}).

However, Type-1 (T1) Fuzzy sets have limited capabilities to directly handle data uncertainties, where \textit{handle} means \textit{to model and minimize the effect of} (\citealt{mendel2007type}). In this context, the performance of type-2 (T2) fuzzy sets can be beyond the T1 (\citealt{karnik1999type}) for highly uncertain situations. A T2 fuzzy set is defined by a lower membership function (LMF) and an upper membership function (UMF), both equivalent to a T1 fuzzy set. There is an area between the LMF and the UMF called footprint of uncertainty - FoU, which is what increases interpretability of T2 fuzzy sets as the degree of membership of a certain value is considered a fuzzy set while in T1 fuzzy sets it is a single crisp value (\citealt{pinto2022interval}).
An interval type-2 fuzzy set is a simplified version of a type-2 fuzzy set, with the secondary membership grade equal to 1 (\citealt{liang2000interval}). The general type-2 fuzzy sets are a more robust version of T2 FSs, which can handle more complex and changing systems, and therefore should be better than the interval type-2 fuzzy sets to deal with uncertainties (\citealt{coupland2007geometric}). Type-2 fuzzy logic has attracted attention due to its wide range of applications, such as transport scheduling (\citealt{john1996type}), forecasting of time-series (\citealt{karnik1999applications}), signal processing (\citealt{mendel1997designing}), pattern recognition (\citealt{mitchell2005pattern}), decision making (\citealt{yager1980fuzzy}), and speech recognition (\citealt{zeng2006type}).

As stated by \citet{lughofer2011evolving}, one of the main challenges in the eFS field is the drift and shift handling in data streams. In these streams, the underlying data-generating process changes (drifts) over time. Incremental learning is one of the common approaches for training and testing incoming data streams; this process handles the noise and the concept drift with the incoming data. Noise and drift in data represent the change in the relationship between the input vector and the output(s). The eFS class uses the evolving clustering to detect the concept drift and then update the rule antecedent and consequent along with managing the number of rules. However, clustering is susceptible to outliers and noise (\citealt{varshney2023literature}).

In this paper, we propose a new evolving fuzzy system, the enhanced evolving Participatory Learning Kernel Recursive Least Squares algorithm based on Fuzzy Sets Measures (ePL-KRLS-FSM+). This model allows us to create type-2 fuzzy sets by introducing a secondary membership function (MF), and measures them using type-2 fuzzy metrics, to evolve the model's structure and compute the results. This secondary MF is beneficial as it enables the definition of membership values that are noisy or uncertain by using a less precise representation.

This paper is an extended version of \citet{de2022kernel, marques2022kernel}, the evolving Participatory Learning Kernel Recursive Least Squares algorithm based on Fuzzy Sets Measures (ePL-KRLS-FSM). The new model proposes different methods to calculate the compatibility measure of the forecasting model; and new ways for designing the fuzzy sets, by effectively using the \textit{fuzzycreator} (\citealt{mcculloch2017fuzzycreator}) as a data-driven tool for generating T1 FSs, allowing the model to better handle inaccuracies within the data. The innovations of the ePL-KRLS-FSM+ are presented below:

\begin{itemize}
    \item The proposed system introduces new ways for generating fuzzy sets, such as Singleton and Interval Polling, and Discrete FSs (presented in Subsection \ref{GFS}).
    \item The system allows to change the parameters for fuzzy sets, designing new forms to analyze the structure of the FSs (presented in Subsection \ref{FS}).
    \item Methods for creating type-2 fuzzy sets are implemented in the extended work, increasing the interpretability of the models (presented in Section \ref{Model}).
    \item New possibilities for comparing data sets in the compatibility measure have been added; being type-1, entropy and inclusion measures (presented in Subsection \ref{T1FM}).
\end{itemize}

These preliminary findings allowed us to keep exploring the model and investigate different techniques in order to improve its performance when dealing with chaotic forecasting scenarios. In this context, this paper's main contributions are summarized as follows: 

\begin{itemize}
    \item We provide detailed information about the ePL-KRLS-FSM+ architecture, which is an extension of the ePL-KRLS-FSM, previously proposed in-brief in \citet{de2022kernel, marques2022kernel}.
    \item We conduct a performance analysis of the model by using two complex datasets: the chaotic time series Mackey-Glass delay differential equation (\citealt{mackey1977oscillation}) with different degrees of chaos, and the main stock index of the Taiwan Capitalization Weighted Stock Index - TAIEX (\citealt{TAIEX}).
    \item Model performance is compared to other eFS models and classical approaches and is analysed in terms of error metrics, runtime and the number of final rules. 

\end{itemize}

\noindent The major findings of this paper are:

\begin{itemize}
    \item ePL-KRLS-FSM+ achieved accurate results, presenting the lowest error metrics suggesting that the forecasting model can predict chaotic time series with high accuracy. 
    \item ePL-KRLS-FSM+ presented the lowest number of final rules in all experiments, indicating low computational complexity.
    \item The possibility of implementing different compatibility measures makes the model robust for various applications, changing the parameters as needed.
    
\end{itemize}

The structure of this paper is organized as follows: Section \ref{Background} presents an overview of the ePL-KRLS-FSM+, the evolving Participatory Learning, the KRLS methodology, and the generation of fuzzy sets. Section \ref{Model} describes the proposed improvement, i.e., the creation of type-2 fuzzy sets. Section \ref{Results} comprises the experimental results. And finally, Section \ref{Conclusion} concludes the paper.

%% Background
\section{Background}\label{Background}

This section presents related work that has been developed on evolving Fuzzy Systems and then introduces the main concepts that drive the proposed model.

%% Related work about evolving Fuzzy Systems
\subsection{Related work about evolving Fuzzy Systems}

The evolving fuzzy rule-based model was introduced by \citet{angelov2001evolving}, this system has shown a notorious advantage compared to pure neural networks models, presenting evident and explainable results (\citealt{angelov2010evolving, leite2020overview}). \citet{rong2006sequential} developed the Sequential Adaptive Fuzzy Inference System that uses an extended Kalman filter to update the fuzzy rules (SAFIS). To overcome the limitations of SAFIS such as the complexity of the rule's influence and the difficulty to cope with high-dimensional input spaces, \citet{rong2011extended} proposed the extended version of SAFIS (ESAFIS). \citet{lughofer2008flexfis} created the Flexible Fuzzy Inference Systems (FLEXFIS), an algorithm that uses a linear polynomial to exploit the Takagi–Sugeno model, this model demands low computational cost and is adequate for online operations (\citealt{trawinski2011investigation}). \citet{lughofer2010sparsefis} presented a model that optimizes the consequent parameters and sparses out the unimportant rules, the Sparse Fuzzy Inference Systems (SparseFIS), using a numerical optimization mechanism to define a compact ruleset (\citealt{serdio2013data}).

\citet{vskrjanc2015evolving} implemented a model to deal with highly noisy data, the evolving Gustafson–Kessel Possibilistic c-Means clustering (eGKPCM). \cite{maciel2016evolving} developed the evolving Possibilistic Fuzzy Modeling approach (ePFM), which uses the Gustafson–Kessel (GK) clustering algorithm to identify clusters with different shapes and orientations while processing the data recursively. Also, this model uses utility measure to evaluate the quality of the current cluster structure. These implementations make the model robust to modeling volatility dynamics and nonlinear volatility forecasting with jumps.

\citet{ge2018self} detailed a model with self-learning parameters, the Self-evolving Fuzzy System (SEFS), this method improves the model’s accuracy by auto-adjusting the rule’s adding speed, reducing underfitting or overfitting. \citet{leite2019optimal} created the evolving Optimal Granular System (eOGS), making it possible to trade off multiple objectives. \citet{alves2020enhanced} implemented the Set-Membership (SM) and the Enhanced Set-Membership (ESM) in the ePL-KRLS (\citealt{alves2021novel}) model, the objective of the membership is to update the rate of change of the arousal index and control the rule creation speed, enhancing the performance of the models. Finally, \cite{ge2020learning} presented a model with self-learning thresholds, the evolving Fuzzy System Self-learning/adaptive Thresholds (EFS-SLAT). 

%% An overview of ePL-KRLS-FSM+
\subsection{An overview of ePL-KRLS-FSM+}

The ePL-KRLS-FSM+ is a ruled-based system that uses the Tagaki-Sugeno (TS) inference method (\citealt{takagi1985fuzzy}). The TS consists of two main parts, the antecedent and the consequent, as following described:

\begin{equation}
    \Re_{i}: \quad \mbox{IF} \quad \underbrace{x \quad \mbox{is} \quad  \mathcal{A}_{i}}_\text{Antecedent} \quad \mbox{THEN} \quad \underbrace{y_{i} = f_{i}(x, \theta_{i})}_\text{Consequent}
\end{equation}

\noindent where $\Re_{i}$ is the $i$th fuzzy rule, $i = 1, 2, \ldots, R$ is the number of fuzzy rules, $x = [x_{1}, \ldots, x_{m}]^{T} \in \mathbb{R}^{m}$ is the input, $m$ is the number of attributes in the input vector, $\mathcal{A}_{i}$ is the fuzzy set (FS) of the $i$th fuzzy rule, and $y_{i}$ is the output of the $i$th rule calculated as a function of the input and the consequent parameters. 

The FSM model uses the Participatory Learning concept in the antecedent part and the Kernel Recursive Least Square in the consequent, being defined as follows:

\noindent \textbf{Antecedent:} Consists of clusters formed from the input space, which are the rules. Each rule is composed of similar inputs.

\noindent \textbf{Consequent:} Consists of a set of parameters associated with a rule, which are the consequent parameters.

%% Participatory Learning
\subsection{Participatory Learning}

Participatory Learning (PL) is a continuous process that works by revising the concepts in the system to create new data, where this new data impacts the self-organization or/and compatibility of the current knowledge (\citealt{lima2010evolving}). The actual knowledge is formed by fuzzy rules, and each rule is characterized by a center, which is an estimation of the rule’s mean. When a new input enters the system, the model compares the new vector with the center of all created rules (\citealt{de2022kernel, marques2022kernel}). If this input is similar enough to the most similar cluster data, the model includes the new input into that rule and updates the cluster center. Otherwise, a new rule is created (\citealt{lemos2010multivariable}).

The center of a new rule is initialized with the input vector, i.e., $\upsilon_{i}^{k} = x^{k}$, where $\upsilon_{i}^{k}$ is the center of the $i$th rule at the $k$th iteration, $x^{k} = [x_{1}, \ldots, x_{m}]^{T}$ is the input vector of attributes with $m$ elements. When a rule receives a new input vector, the cluster center is recursively updated according to Equation \eqref{nu}.

\begin{equation} \label{nu}
    \upsilon_{i}^{k} = \upsilon_{i}^{k - 1} + \alpha (c_{i}^{k})^{(1 - a_{i}^{k})} (x^{k} - \upsilon_{i}^{k - 1})
\end{equation}

\noindent where $\alpha \in [0, 1]$ is the learning rate.

In this model, the rules are defined using the compatibility measure and the arousal index. The compatibility measure indicates the degree of similarity between the new input and the cluster center, where 0 indicates no similarity between the rule and the input, and 1 represents the maximum similarity.

The arousal index works as a complement of the compatibility measure, reducing the effect of outliers and indicating the necessity to create a new rule (\citealt{alves2021novel}), and is given as $a_{i}^{k} \in [0, 1]$.  If the smallest arousal index is greater than a threshold, i.e., $a_{i}^{k} > \tau$, where $i = \arg \min_{i} \{a_{i}^{k}\}$ and $\tau = \beta$ (the same as (\citealt{alves2020enhanced}), the model creates a new rule. Otherwise, the input vector is included in the rule with the highest compatibility measure. The arousal index is calculated as:

\begin{equation}
    a_{i}^{k} = a_{i}^{k - 1} + \beta (1 - c_{i}^{k} - a_{i}^{k - 1})
\end{equation}

\noindent where $\beta \in [0, 1]$ is a parameter that controls the growth rate of $a_{i}^{k}$.

To remove underused rules in the system, the utility measure is implemented, avoiding overfitting noisy data. The pruning mechanism eliminates complex rules with low coverage that contain irrelevant literals, which have only been created to enclose noisy examples (\citealt{furnkranz1997pruning}). The model eliminates the rule when a utility of a rule gets lower than a threshold ($U_{i}^{k} < \epsilon$), following the equation:

\begin{equation}
    U_{i}^{k} = \dfrac{\Sigma_{l = 1}^{k} \lambda_{i}^{l}}{k - I_{i}}
\end{equation}

\begin{equation}
    \lambda_{i}^{l} = \dfrac{\tau_{i}}{\Sigma_{rule = 1}^{R^k} \tau_{rule}}
\end{equation}

\begin{equation}
    \tau_{i} = \mu_{i1} \times \mu_{i1} \times \cdots \mu_{ij}
\end{equation}

\begin{equation} \label{mu-ij}
    \mu_{ij} = \dfrac{e^{- \| x^{k} - \upsilon_{ij}^{k} \|}}{2 \sigma^{2}}
\end{equation}

\noindent where $\lambda_{i}^{l}$ is the normalized activation level of the $i$th rule at the $l$th iteration, $k$th is the current iteration, and $I_{i}$ is the iteration when the $i$th rule was created, $\tau_{i}$ is the activation level of the $i$th rule, $R^{k}$ is the number of rules at the current iteration, $\mu_{ij}$ is calculated according to the Equation \eqref{mu-ij}, $\upsilon_{ij}^{k}$ is the $j$th element of the $i$th rule center, i.e., $j = 1, 2, \ldots ,m$, and $\sigma$ defines the spread of the antecedent part.

%% Kernel Recursive Least Square algorithm
\subsection{Kernel Recursive Least Square algorithm}

The kernel recursive least square (KRLS) (\citealt{engel2004kernel}) is a method that implements a sparsification procedure to achieve lower computational costs and is closely related to \citet{csato2001sparse} in the context of learning with Gaussian Processes (\citealt{williams1998prediction}).

The KRLS estimates the consequent parameters using a collection of past inputs. Each collection forms the local dictionary, i.e., $\mathcal{D}_{i}^{k} = [d_{i1}, \ldots, d_{in_i}]$, where $\mathcal{D}_{i}^{k}$ is the local dictionary of the $i$th rule at the $k$th iteration and $n_{i}$ is the number of input vectors stored in the dictionary. Only relevant input vectors are stored in the local dictionaries, where the inputs' importance is based on their distance to the nearest element, as shown in Equation \eqref{dis-x}.

\begin{equation} \label{dis-x}
    dis_{x} = \mathop{\min}_{\forall d_{ij} \in \mathcal{D}_{i}^{k}} \| x^{k} - d_{ij}^{k} \|
\end{equation}

\noindent where $d_{ij} = [d_{1}, \ldots , d_{m}]^{T}$ is the $j$th input vector of the dictionary. If the distance is greater than a threshold that defines the inclusion of new inputs into the dictionaries, the model adds the input vector into the local dictionary, as follows:

\begin{equation}
    \mbox{IF} \quad dis_{x} \geq 0.1 \nu_{ij}^{k} \quad \mbox{THEN} \quad \mathcal{D}_{i}^{k} = [\mathcal{D}_{i}^{k - 1} \cup x^{k}]
\end{equation}

\noindent where $j = \arg \min_{j} \| x^{k} - d_{ij}^{k} \|$, $\nu_{i}^{k}$ is the kernel size of the dictionary, and $n_i$ is the number of input vectors in the local dictionary. When the model adds a vector to the dictionary, the consequent parameters are updated according to the following equations.

\begin{equation} \label{CP}
    \theta_{i}^{k} = \left[ \begin{array}{c} \theta_{i}^{k - 1} - z_{i}^{k} [r_{i}^{k}]^{-1} \hat{e}^{k} \\ {[r_{i}^{k}]}^{-1} \hat{e}^{k} \end{array} \right]
\end{equation}

\begin{equation} \label{UP}
    Q_{i}^{k} = (r^{k})^{-1} \left[ \begin{array}{cc} Q_{i}^{k - 1} r^{k} + z^{k} (z^{k})^{T} & - z^{k} \\ - (z^{k})^{T} & 1 \end{array} \right]
\end{equation}

\begin{equation} \label{g}
    g^{k} = [\kappa \langle d_{ij}^{k} , x^{k} \rangle, \ldots , \kappa \langle d_{in_i}^{k} , x^{k} \rangle]^{T}
\end{equation}

\begin{equation} \label{kappa}
    \kappa \langle x^{i}, x^{j} \rangle = \exp \left( - \dfrac{\| x^{i} - x^{j} \|^{2}}{2 \sigma^{2}} \right)
\end{equation}

\noindent where $\theta_{i}^{k - 1} = [\theta_{i1}^{k - 1}, \ldots, \theta_{in_1}^{k - 1}]^{T}$, $z^{k} = Q_{i}^{k - 1} g^{k}$, $r^{k} = \lambda + \kappa \langle x^{k}, x^{k} \rangle - (z^{k})^{T} g^{k}$, $\hat{e}^{k} = y^{k} - g^{k} \theta_{i}^{k - 1}$ is the error estimator, $\kappa \langle \ldots \rangle$ is the Gaussian-Kernel function and $\sigma$ is the kernel width that controls the linearity of the model. \citet{fan2016kernel} suggested initial values for the kernel width as $0.2 \leq \sigma \leq 0.5$. If the input vector is included in the dictionary, the matrix P is updated as:

\begin{equation} \label{P}
    P_{i}^{k} = \left[ \begin{array}{cc} P_{i}^{k - 1} & 0 \\ 0^{T} & 1 \end{array} \right]
\end{equation}

\noindent where $P_{i}^{k}$ is initialized with one. Otherwise, if the input is not included, the consequent parameters are updated in Equation \eqref{theta}, and the matrix P in Equation \eqref{P eq}.

\begin{equation} \label{theta}
    \theta_{i}^{k} = \theta_{i}^{k - 1} + Q_{i}^{k} q_{i}^{k} \hat{e}^{k}
\end{equation}

\begin{equation} \label{Q}
    Q_{i}^{k} = Q_{i}^{k - 1}
\end{equation}

\begin{equation} \label{q}
    q_{i}^{k} = \dfrac{P_{i}^{k - 1} z^{k}}{1 + (z^{k})^{T} P_{i}^{k - 1} (z^{k})^{T}}
\end{equation}

\begin{equation} \label{P eq}
    P_{i}^{k} = P_{i}^{k - 1} - \dfrac{P_{i}^{k - 1} z^{k} (z^{k})^{T} P_{i}^{k - 1}}{1 + (z^{k})^{T} P_{i}^{k - 1} (z^{k})^{T}}
\end{equation}

The consequent parameter of a new rule is initialized as follows:

\begin{equation}
    \theta_{i}^{k} = [\lambda + \kappa \langle x^{k}, x^{k} \rangle]^{- 1} y^{k}
\end{equation}

\noindent where $\lambda \in [0, 1]$ is a parameter of regularization and $y^{k}$ is the desired output. The model computes the output using the most compatible rule, i.e., $\hat{y} = \hat{y}_{i} \vert i = \arg \max_{i} \{c_{i}^{k} \}$, being calculated according to Equation \eqref{output}.

\begin{equation} \label{output}
    \hat{y} = \sum_{j = 1}^{n_i} \theta_{ij}^{k} \times \kappa \langle x^{k}, d_{ij}^{k} \rangle
\end{equation}

If the model updates an existing rule, the kernel size is updated as Equation \eqref{KSU}.

\begin{scriptsize}
\begin{equation} \label{KSU}
    \nu_{i}^{k} = \sqrt{(\nu_{i}^{k - 1}) + \dfrac{\| x^{k} - \upsilon_{i}^{k} \| - (\nu_{i}^{k - 1})^{2}}{N_{i}^{k}} + \dfrac{ (N_{i}^{k - 1})^{2} - \| \upsilon_{i}^{k} - \upsilon_{i}^{k - 1} \|^{2}}{N_{i}^{k}}}
\end{equation}
\end{scriptsize}

\noindent where $N_{i}^{k}$ is the number of inputs in the $i$th rule at the $k$th iteration. If a new rule is created, the model initializes the kernel size of the rule as follows: 

\begin{equation} \label{KSR}
    \nu_{R + 1}^{k} = \dfrac{\| x^{k} - \upsilon_{i}^{k} \|}{\sqrt{2 \log (\eta_{\max})}}
\end{equation}

\begin{equation} \label{eta}
    \eta_{k} = e^{- 0.5} \left( \dfrac{2}{1 + e^{- \hat{e}^{k}}} - 1 \right)
\end{equation}

\noindent where $\eta_{\max}$ is the maximum value of $\eta_{k}$ , for $K = 1, 2, \ldots , k$, $k$ is the current iteration, $\eta_{k}$ is calculated recursively using Equation \eqref{eta}, $\hat{e}^{k} = 0.8 \hat{e}^{k - 1} + \| y^{k} - \hat{y}^{k} \|$ and $\hat{e}^{1} = 0$. In this model, the kernel size is updated based on the error. Higher error values mean that more input data is stored in the dictionary, improving model’s performance (\citealt{alves2021novel}).

%% Generation of fuzzy sets
\subsection{Generation of fuzzy sets} \label{GFS}

For the proposed model we consider four methods of generating fuzzy sets directly from data, as follows:

\noindent \textbf{Gaussian Function:} Calculates the membership values based on mean and standard deviation of the data.

\noindent \textbf{Singleton Polling:} The membership values are generated using a histogram-based approach, taking a list of numerical values within a given range and, producing discrete FSs.

\noindent \textbf{Interval Polling:} The memberships are calculated using a method called Interval Agreement Approach (IAA), creating FSs from interval-valued data (\citealt{wagner2014interval}). This method only works with interval data.

\noindent \textbf{Discrete Fuzzy Sets:} The discrete FSs are generated with no interpolation, where the memberships are explicitly stated for any value. Only discrete data works in this method.

%% Gaussian Functions
\subsubsection{Gaussian Functions}

The Gaussian MFs are generated by calculating the mean and the standard deviation of the input vector and the cluster center, creating the fuzzy sets. The membership values are calculated as follows:

\begin{equation} \label{MF}
    \mu_{C}(Y^{k}) = e^{-0,5} \cdot \left(\dfrac{Y^{k} - \mathcal{M}^{k}}{\mathcal{S}^{k}} \right)^2
\end{equation}

\noindent where $\mu_{C}(Y^{k})$ is the membership function, $Y^{k} = [Y_{1}^{k}, \ldots , Y_{m}^{k}]^{T} \in \mathbb{R}^{m}$ is the actual value at the $k$th iteration, $\mathcal{M}^{k}$ is the mean and $\mathcal{S}^{k}$ is the standard deviation. As the Equation \eqref{MF} is used both for the input vector and the cluster center, the notation of the terms ($\mu_{C}(Y^{k}), Y^{k}, \mathcal{M}^{k}, \mathcal{S}^{k}$) helps to prevent redundancy.

If the point of the actual data is lower or higher than the endpoint of the function, the membership is zero, as follows:

\begin{equation} \label{Y min max}
    \mathcal{P}^{k} = \left\{\begin{array}{l}
    Y_{\min}^{k} = \mathcal{M}^{k} - (4 \mathcal{S}^{k}) \\
    Y_{\max}^{k} = \mathcal{M}^{k} + (4 \mathcal{S}^{k}) \end{array} \right.
\end{equation}

\begin{equation}\label{MF 0}
    \begin{aligned}
    &\mbox{IF} \quad Y_{p}^{k} < Y_{\min}^{k} \quad  \mbox{OR} \quad Y_{p}^{k} > Y_{\max}^{k}\\
    &\mbox{THEN} \quad \mu_{C}(Y) = 0
   \end{aligned}
\end{equation}

\noindent where $Y_{p}^{k} = [Y_{p,1}^{k}, \ldots , Y_{p,m}^{k}]^{T} \in \mathbb{R}^{m}$ is the actual value of the point at the $k$th iteration. The notation of the term $Y_{p}^{k}$ helps to prevent redundancy for the input vector and the cluster center.

%% Singleton Polling
\subsubsection{Singleton Polling}

To generate the Singleton Polling MFs, the model calculates the FSs left and right endpoints. The membership values are calculated as follows:

\begin{footnotesize}
\begin{equation} \label{MFP}
    \mu_{C}(Y_{p}^{k}) = \left\{ \begin{array}{lcl}
    
    \mathcal{I}^{k} * (X^{k} - X_{L}^{k}) + Y_{p,L}^{k} , & \mbox{if}
    & I = Linear \\[0.1cm] 

    \mathcal{L}(X^{k}, Y_{p}^{k})  , & \mbox{if} & I = Lagrange
    
    \end{array}\right.
\end{equation}
\end{footnotesize}

\noindent where $\mathcal{I}^{k} = \frac{(Y_{p,R}^{k} - Y_{p,L}^{k})}{(X_{R}^{k} - X_{L}^{k})}$ is the slope, $\mathcal{L}(X^{k}, Y_{p}^{k})$, is the Lagrange function, $I$ is the interpolation method, $X^{k} = [X_{1}^{k}, \ldots , X_{m}^{k}]^{T} \in \mathbb{R}^{m}$ is the actual point at the $k$th iteration, $X_{L}^{k}$ is the antecedent bisection between the sorted point and the actual point at the $k$th iteration ($X_{R}^{k}$ is the actual bisection), and the $Y_{p,L}^{k}$ is the antecedent value of the bisection ($Y_{p,R}^{k}$ is the actual bisection). The notation of the terms ($\mathcal{I}^{k}, X^{k}, X_{L}^{k}, X_{R}^{k}$) helps to prevent redundancy.

%% Interval Polling
\subsubsection{Interval Polling}

To generate the Interval Polling MFs, the model calculates the actual value of the points of the FSs. The membership values are calculated as follows:

\begin{equation} \label{MFI}
    \mu_{C}(Y_{\mathcal{I}}^{k}) = \left\{ \begin{array}{lcl}
    
    \frac{Y_{\mathcal{I}}^{k}}{Y_{\mathcal{I},\max}^{k}} , & \mbox{if}
    & N = True \\[0.35cm] 

    \frac{Y_{\mathcal{I}}^{k}}{\mathcal{T}^{k}}  , & \mbox{if} & N = False
    
    \end{array}\right.
\end{equation}

\noindent where $Y_{\mathcal{I}}^{k} = [Y_{\mathcal{I},1}^{k}, \ldots , Y_{\mathcal{I},m}^{k}]^{T} \in \mathbb{R}^{m}$ is the actual value of the interval at the $k$th iteration, $\mathcal{I} \in [0, m]$ is the interval of the actual value, $Y_{\mathcal{I},\max}^{k}$ is the largest value of the actual value of the interval at the $k$th iteration, $N$ is the normalization, and $\mathcal{T}^{k}$ is the number of total intervals of the actual value of the interval at the $k$th iteration. The notation of the term $\mathcal{T}^{k}$ helps to prevent redundancy.

\subsubsection{Discrete Fuzzy Sets}

To generate the Discrete MFs, the model calculates the actual value of the points of the FSs. The membership values are calculated as follows:

\begin{equation} \label{MFD}
    \mu_{C}(Y_{p}^{k}) = Y_{p}^{k}
\end{equation}

\noindent where $Y_{p}^{k} = [Y_{p,1}^{k}, \ldots , Y_{p,m}^{k}]^{T} \in \mathbb{R}^{m}$ is the actual value of the point at the $k$th iteration. The notation of the term $Y_{p}^{k}$ helps to prevent redundancy.

%% Fuzzy sets analysis
\subsection{Fuzzy sets analysis}\label{FS}

To analyze the FSs, this model uses metrics to compare the affinity and the points between the data. These analyses are based on similarity and distance measures (\citealt{mcculloch2017fuzzycreator}), as following explained:

\noindent \textbf{Similarity:} Determines the degree of resemblance of two fuzzy sets, which is given as $s(A, B) \in [0, 1]$, where 0 indicates no resemblance, and 1 indicates identical fuzzy sets (\citealt{cross2002similarity}).

\noindent \textbf{Distance:} Determines how far two fuzzy sets are placed within a universe of discourse, which is given as $d(A, B) \in \mathbb{R}^{+}$, where higher values indicate how further the fuzzy sets are from each other (\citealt{blanco2014distance}).

In order to have a better analysis of the FSs, the ePL-KRLS-FSM+ allows to set and change the following fuzzy sets' parameters: normalization, the universe of discourse, and the number of discretizations. This change can be made for all FSs or individually.

The FSs generated from data can be normal or non-normal, the option for normalization is $True$ by default. The universe of discourse (UOD) is given as $[0; 10]$, and the number of discretizations is given as 101 by default.

%% Proposed Model
\section{The concept of evolving Participatory Learning Kernel Recursive Least Squares algorithm based on Type-2 Fuzzy Sets Measures}\label{Model}

The ePL-KRLS-FSM+ is a robust fuzzy rule-based system, that allows changing the metrics in the compatibility measure and the MFs of the fuzzy sets. These changes allow to capture the uncertainty of data through transformation and relative comparisons (\citealt{mcculloch2017fuzzycreator}). The methods for creating T2 FSs are: Interval Agreement Approach with Interval Data, Gaussian, and Polling with Singleton Data.

The data is split into two subsets, each subset is constructed into a type-1 fuzzy set, and they are aggregated into a type-2 fuzzy set, forming the secondary membership function (\citealt{mcculloch2017fuzzycreator}). These T2 FSs are compared using similarity and distance measures for General and/or Interval data (\citealt{mcculloch2018measuring}). The subsets of the input vector and cluster center are defined as follows:

\begin{equation} \label{subset}
\begin{matrix}
    {}_{1}{Y}^{k} = [Y_{1}^{k}, \ldots , Y_{\frac{m}{2}}^{k}]^{T} \in \mathbb{R}^{m} \\[0.4cm]

    {}_{2}{Y}^{k} = [Y_{(\frac{m}{2} + 1)}^{k}, \ldots , Y_{m}^{k}]^{T} \in \mathbb{R}^{m}
\end{matrix}
\end{equation}

\noindent where ${}_{1}Y^{k}$ and ${}_{2}Y^{k}$ are the notation of the input vector and the cluster center at the \textit{k}th iteration, respectively, and $m$ is the number of attributes. If $m$ is an odd value, the subsets are defined as:

\begin{equation} \label{subset-odd}
\begin{matrix}
    {}_{1}{Y}^{k} = [Y_{1}^{k}, \ldots , Y_{(\frac{m}{2} + \frac{1}{2})}^{k}]^{T} \in \mathbb{R}^{m} \\[0.4cm]

    {}_{2}{Y}^{k} = [Y_{(\frac{m}{2} + \frac{1}{2})}^{k}, \ldots , Y_{m}^{k}]^{T} \in \mathbb{R}^{m}
\end{matrix}
\end{equation}

The membership values of each subset are calculated according to Section \ref{GFS}, forming type-1 fuzzy sets. Afterwards, they are aggregated into a type-2 fuzzy set, as follows:

\begin{equation} \label{MFT2}
    \mu_{\tilde{C}}(Y^{k}) = ({}_{1}{Y}^{k}, \ {}_{2}{Y}^{k})
\end{equation}

The T2 FSs are analyzed through similarity and distance measures based on interval type-2 (IT2) and/or general type-2 (GT2). Basic calculations can be performed on IT2 and GT2, including calculating the primary membership for a given value and calculating the alpha-cuts of the lower and upper MFs. They are defined as:

\noindent \textbf{Interval Type-2 Fuzzy Sets:} Defined by two MFs of the same class (e.g. both Gaussian). The type-reduction is achieved with the Karnik-Mendel centre-of-sets method (\citealt{karnik2001centroid}).

\noindent \textbf{General Type-2 Fuzzy Sets:} Constructed using the zSlices/alpha-plane representation (\citealt{wagner2010toward}), in which the secondary MF has the value 1 at the centre of the footprint of uncertainty (FOU) and the membership decreases linearly towards the edge of the FOU.

The compatibility measure for T2 fuzzy sets is calculated as:

\newpage

\begin{strip}
\begin{equation} \label{CMT2}
    c_{i}^{k} = \left\{ \begin{array}{lcl}

    1 - \dfrac{1}{2n} \sum\limits_{k=1}^{n} ( \vert \underline{\mu}_{\tilde{A}}(x^k) - \underline{\mu}_{\tilde{B}}(\upsilon_i^k) \vert + \vert \overline{\mu}_{\tilde{A}}(x^k) - \overline{\mu}_{\tilde{B}}(\upsilon_i^k) \vert) , & \quad \mbox{if} \quad 
    & M = zeng\_li \\[0.8cm]
    \dfrac{\sum_{k=1}^{\zeta} z_k \ s_j^{IT2} (\tilde{A}_{z_k}, \tilde{B}_{z_k})}{\sum_{i=1}^{\zeta} z_k} , & \mbox{if}
    & M = jaccard\_gt2 \\[0.8cm]

    \dfrac{1}{\Delta + 1} \sum\limits_{\alpha = 0, \frac{1}{\Delta}, \frac{2}{\Delta}, \cdots, \frac{\Delta - 1}{\Delta}, 1} s_j^{IT2} (\tilde{A}_{z_{\alpha}}, \tilde{B}_{z_{\alpha}}) , & \mbox{if}
    & M = zhao\_crisp \\[0.8cm]

    \bigcup_{\forall z_k} \dfrac{z_k}{s_j^{IT2} (\tilde{A}_{z_k}, \tilde{B}_{z_k})}, & \mbox{if}
    & M = hao\_crisp \\[0.8cm]

    \dfrac{1}{m} \sum\limits_{\chi \in X} \dfrac{\sum_{u \in J_{\chi}} \min\{u \cdot f_{\chi} (u), u \cdot g_{\chi} (u)\}}{\sum_{u \in J_{\chi}} \max\{u \cdot f_{\chi} (u), u \cdot g_{\chi} (u)\}} , & \mbox{if}
    & M = yang\_lin \\[0.8cm]
    
    \dfrac{1}{m} \sum\limits_{\chi \in X} \dfrac{\min \{\sum_{u \in J_{\chi}} 1 - u \cdot f_{\chi} (u), \sum_{u \in J_{\chi}} 1 - u \cdot g_{\chi} (u)\}}{\max \{\sum_{u \in J_{\chi}} 1 - u \cdot f_{\chi} (u), \sum_{u \in J_{\chi}} 1 - u \cdot g_{\chi} (u)\}} , & \mbox{if}
    & M = mohamed\_abdaala \\[0.8cm]

    1 - \dfrac{\sum_{k=1}^n H_f (\tilde{A}(x^k), \tilde{B}(\upsilon_i^k))}{n} , & \mbox{if}
    & M = hung\_yang
    
    \end{array}\right.
\end{equation}
\end{strip}

\noindent where $n$ is the number of discretizations of data, $\tilde{A}$ is the T2 FS of the input vector, $\tilde{B}$ is the T2 FS of the cluster center, $\zeta$ is the number os number of zSlices, $z_k$ is the actual zSlice at the $k$th iteration, $z_{\alpha}$ is the actual zSlice at the actual $\alpha$-cut, $s_j^{IT2}$ is the similarity measure for IT2 fuzzy sets, being called \textit{jaccard\_it2} (\citealt{wu2008vector, nguyen2008computing}), $m$ is the number of attributes, $X$ is the universe of discourse, $J_{\chi}$ is the primary membership of $\chi$ ($J_{\chi} \subseteq [0, 1]$), $f_{\chi}(u)$ and $g_{\chi}(u)$ are the secondary membership functions for $\mu_{\tilde{A}}({}_{1}x^{k}, {}_{2}x^{k})$ and $\mu_{\tilde{B}}({}_{1}\upsilon_i^{k}, {}_{2}\upsilon_i^{k})$, respectively. The $s_j^{IT2}$ is determined as follows:

\begin{equation}
\begin{array}{lc}
    s_j^{IT2} (\tilde{A}, \tilde{B}) = & \dfrac{\sum_{k=1}^{n} \min(\overline{\mu}_{\tilde{A}}(x^k), \overline{\mu}_{\tilde{B}}(\upsilon_i^k))}{\sum_{k=1}^{n} \max(\overline{\mu}_{\tilde{A}}(x^k), \overline{\mu}_{\tilde{B}}(\upsilon_i^k))} \ + \\[0.6cm]
    
    & \dfrac{\sum_{k=1}^{n} \min(\underline{\mu}_{\tilde{A}}(x^k), \underline{\mu}_{\tilde{B}}(\upsilon_i^k))}{\sum_{k=1}^{n} \max(\underline{\mu}_{\tilde{A}}(x^k), \underline{\mu}_{\tilde{B}}(\upsilon_i^k))}
\end{array}
\end{equation}

The $H_f$ is calculated as:

\begin{equation}
    H_f = \dfrac{\sum_{k=1}^{\iota} \alpha_k H (\tilde{A}_{\alpha_k}, \tilde{B}_{\alpha_k})}{\sum_{k=1}^{\iota} \alpha_k}
\end{equation}

\noindent where $H (\tilde{A}_{\alpha_k}, \tilde{B}_{\alpha_k}) = max(L(\tilde{A}_{\alpha_k}, \tilde{B}_{\alpha_k}), L(\tilde{B}_{\alpha_k}, \tilde{A}_{\alpha_k}))$ and $L (U, V) = \inf \{\lambda \in [0, \infty] \ \vert \ U^{\lambda} \supset V \}$, the notation of $U$ and $V$ helps to prevent redundancy.

%% Measures for type-2 fuzzy sets
\subsection{Measures for type-2 fuzzy sets} \label{T2FM}

The equations in the compatibility measure for T2 fuzzy sets are similarity metrics proposed by \citet{zeng2006relationship}, \citet{mcculloch2013extending}, \citet{zhao2014new}, \citet{hao2014similarity}, \citet{lin2007similarity, yang2009similarity}, \citet{mohamed2011applying}, and \citet{hung2004similarity}. The model is called ePL-KRLS-T2FSM\_zl for \textit{zeng\_li}, ePL-KRLS-T2FSM\_jg2 for \textit{jaccard\_gt2}, ePL-KRLS-T2FSM\_zc for \textit{zhao\_crisp}, ePL-KRLS-T2FSM\_hc for \textit{hao\_crisp}, ePL-KRLS-T2FSM\_yl for \textit{yang\_lin}, ePL-KRLS-T2FSM\_ma for \textit{mohamed\_abdaala}, and ePL-KRLS-T2FSM\_hy for \textit{hung\_yang} measure.

The measure proposed by \citet{zeng2006relationship} introduces the concept of entropy of interval valued fuzzy set different from \citet{burillo1996entropy}, giving a method to describe entropy of interval valued fuzzy set based on its similarity measure. Similarity measure and entropy of interval valued fuzzy sets can be transformed by each other based on their axiomatic definitions. These contributions can be applied in many fields such as pattern recognition, image processing, and fuzzy reasoning (\citealt{zeng2006relationship}).

The ePL-KRLS-T2FSM\_jg2 support the idea that a measure of similarity on IT2 FSs can be applied to each zSlice, using zSlices-based GT2 FSs, and the results for each zSlice can be combined. This similarity measure on IT2 is the \textit{jaccard\_it2} proposed by \citet{wu2009comparative} and \citet{nguyen2008computing}.

A new general type-2 fuzzy similarity measure developed by \citet{zhao2014new} is defined by using the $\alpha$-plane theory. The proposed metric improves the shortcomings of the existing measures (e.g. \textit{jaccard\_it2}), being a natural extension of the most popular type-1 fuzzy measures. As it is relative to the numbers of $\alpha$-planes, the larger the numbers of $\alpha$-planes are, more stable the proposed measure is, and the longer the computing times (\citealt{zhao2014new}).

\citet{hao2014similarity} proposed a similarity measure for GT2 FSs using the $\alpha$-plane based on ``zSlice Representation Theorem (RT)'',  this metric is an extension of the Jaccard similarity measure for IT2 FSs (\citealt{wu2009comparative}). This measure calculates the centroid of the representation of similarity as a discrete type-1 fuzzy set, where the FOU influences the GT2 FSs (\citealt{hao2014similarity}).

The \citet{lin2007similarity, yang2009similarity} measure combines the similarity degree between type-2 fuzzy sets with \citet{yang2001cluster} algorithm as a clustering method, obtaining a useful hierarchical tree (according to different $\alpha$-level), being able to cluster the type-2 fuzzy data. In order to consider the FOU of the primary and secondary membership function, this similarity measure is defined (\citealt{lin2007similarity}).

The ePL-KRLS-T2FSM\_ma is similar to the \citet{lin2007similarity, yang2009similarity} measure, applying a new similarity measure to a clustering problem for GT2 fuzzy sets. This method shows more rational clustering results compared with \citet{hung2004similarity}, therefore this model is recommended in conjunction with the \citet{yang2001cluster} algorithm as a clustering method (\citealt{mohamed2011applying}).

Finally, the \textit{hung\_yang} measure combines the Hausdorff and Hamming distance concepts, being extended to subsets of a metric space. This combination helps define the distance for two type-2 fuzzy sets, where the Hausdorff is used to define the secondary membership functions and the Hamming is used to define the primary membership functions (\citealt{hung2004similarity}).

\subsection{Other measures for fuzzy sets} \label{T1FM}

In addition to the similarity and distance metrics, this model allows us to compare them using entropy, which determines how much a fuzzy set is fuzzy (\citealt{de1993definition}) and inclusion, a metric that characterizes the degree of couples of subsets measures (\citealt{sanchez1979inverses}). As the objective of this paper is to evaluate the model's accuracy, these measures will not be used in this work.

In order to further expand the capability of ePL-KRLS-FSM+, a distance measure for T1 fuzzy sets is added to the model, this equation is calculated as follows:

\begin{equation} \label{CMT1}
    \begin{small}
    c_{i}^{k} = \left\{ \begin{array}{lcl}

    \dfrac{\sum_{\alpha \in [0, \lambda]} y_{\alpha} \grave{\bar{d_p}}(A_{\alpha}, B_{\alpha})}{\sum_{\alpha \in [0, \lambda]} y_{\alpha}} , & \mbox{if}
    & M = mcculloch
    
    \end{array}\right.
    \end{small}
\end{equation}

\noindent where $\lambda = \max\{\alpha \ \vert \ A_{\alpha} \neq \emptyset \vee B_{\alpha} \neq \emptyset, \forall \alpha \in (0, 1]\}$, and $\grave{\bar{d_p}}$ is:

\begin{equation}
    \begin{small}
    \grave{\bar{d_p}}(A_{\alpha}, B_{\alpha}) = \left\{ \begin{array}{rcl} 
    \bar{\bar{d_p}}(A_{\alpha}, B_{\alpha}) & \quad & A_{\alpha} \neq \emptyset \wedge B_{\alpha} \neq \emptyset \\ 
    \bar{\bar{d_p}}(A_{\alpha k}, B_{\alpha}) & \quad & A_{\alpha} \neq \emptyset \wedge B_{\alpha} \neq \emptyset \\
    \bar{\bar{d_p}}(A_{\alpha}, B_{\alpha k}) & \quad & A_{\alpha} \neq \emptyset \wedge B_{\alpha} \neq \emptyset
\end{array}\right.
    \end{small}
\end{equation}

\noindent where $A_{\alpha k} = \max\{A_{\alpha} \ \vert \ A_{\alpha} \neq \emptyset, \forall \alpha \in (0, 1]\}$, and $\bar{\bar{d_p}}$ is given as:

\begin{equation}
    \bar{\bar{d_p}}(A_{\alpha}, B_{\alpha}) = \dfrac{1}{\Vert A_{\alpha} \Vert \Vert B_{\alpha} \Vert} \sum_{i=1}^{\Vert A_{\alpha} \Vert} \sum_{j=1}^{\Vert B_{\alpha} \Vert} \bar{d_p}(A_{\alpha}, B_{\alpha})
\end{equation}

This is a directional distance measure where $d(A, B) > 0$ if $A < B$ and
$d(A, B) \geq 0$ if $A \geq B$. Additionally, unlike the above listed metrics, the \textit{mcculloch} measure has no restriction on the normality or convexity of the fuzzy sets. This measure was proposed by \citet{mcculloch2016novel}, being called ePL-KRLS-T1FSM\_mc.

%% Algorithm
\subsection{Algorithm}

The Algorithm \ref{ePL-KRLS-FSM} presents the ePL-KRLS-FSM+ model. The first line starts with the initialization of the cluster center, the dictionary, the kernel size, the matrix $P$, the consequent parameters, and the arousal index. For the training phase, it is started a loop for all input vectors in the second line. 

The FSs are calculated based on Algorithm \ref{fuzzycreator} (line 4). If the method for generation fuzzy sets is type-2, the dataset is divided into subsets (line 6), if the number of attributes is odd, the subsets are created with the center value repeated (line 10), if not, the dataset is equally divided (line 12). If the membership function selected is gaussian, the endpoints of the MF are calculated (line 15) and compared at each point of the data (line 16). If the data value isn't within the minimum and maximum intervals, the MF is zero (line 17). Else, the mean and the standard deviation of the data are used to calculate the membership values  (line 19). If the discrete or singleton polling method is selected, the MFs are calculated based on the actual point of the \textit{k}th iteration (line 23 and 25, respectively). If the method is interval polling, the MFs are calculated in each interval of the actual value (line 30). The outputs are computed for the input vector and the cluster center based on the type of fuzzy set selected, T1 (line 37) or T2 (line 40).

The compatibility measure is calculated based on the measure $M$ adopted (line 5), and the arousal index is defined for all rules (line 6). 

If the smallest arousal index is greater than the value of $\tau$, and no rule was excluded (line 8), the model creates a new rule (line 9), initializes the cluster center, the dictionary and the consequent parameters (line 10), and compute the kernel size (line 11). If $\tau$ is not exceeded by the smallest arousal index, the model looks for the rule with the greater compatibility measure (line 13), and its center is updated. Also, the model calculates the $g$, $z$, and $r$ of the updated rule. 

If all the past inputs in the dictionary are distant enough to the input vector as shown in line 16, the input is added into the dictionary and $Q$, $P$, $\theta$, and $\nu$ are updated (line 19). Otherwise, if the input does not attend the expression presented (line 16), the model updates $Q$, $P$, $\theta$, and $\nu$ as expressed in line 20, and does not include the input into the dictionary. 

Besides, the model computes the utility measure for all rules to eliminate underused rules. When the utility measure is smaller than $\varepsilon$, the model removes this rule according to lines 24, 25, and 26. Finally, the model computes the output using the most compatible rule (line 29). For the test phase, the model executes the commands shown in lines 2, 5, and 29. Table \ref{param-algo} describes the parameters for the algorithms.

\begin{algorithm}
\caption{ePL-KRLS-FSM+ algorithm} \label{ePL-KRLS-FSM}
\begin{algorithmic}[1]
\renewcommand{\algorithmicrequire}{\textbf{Input:}}
\renewcommand{\algorithmicensure}{\textbf{Output:}}

\Require $x, y, \alpha, \beta, \lambda, \tau , \omega, \sigma, \epsilon, A, B, M$
\Ensure  $\hat{y}$

\State \textbf{Initialisation}: $\upsilon_{1}^{1} = x^1, \mathcal{D}_{1}^{1} = x^1, \nu_{1}^{1} = \sigma, P_{1}^{1} = 1, R = 1, \theta_{1}^{1} = [\lambda + \kappa \langle x^1, x^1 \rangle]^{-1}y^{1}, a_{1}^{1} = 0, Excluded\_Rule = False$

\For {$k = 2, 3, \ldots, n$}
    \For {$i = 1, 2, \ldots, R$}
        \State Transform the data in FSs: Algorithm \ref{fuzzycreator}
        \State Compute the compatibility measure $ c_{i}^{k}$: Equation \eqref{CMT1}, \eqref{CMT2}
        \State Compute the arousal index: $a_{i}^{k} = a_{i}^{k - 1} + \beta(1 - c_{i}^{k} - a_{i}^{k - 1})$
    \EndFor
    
    \If {$\arg \min_{i} \{a_{i}^{k}\} > \tau$ \ \mbox{\textbf{and}} \ $Excluded\_Rule == false$}
    
        \State Create a new rule: $R = R + 1$
        \State Initialize $\upsilon_{R}^{k}, \mathcal{D}_{R}^{k}, \theta_{i}^{k} : \upsilon_{R}^{k} = [x^{k}], \mathcal{D}_{R}^{k} = x_{k}, \theta_{i}^{k} = [\lambda + \kappa \langle x^{k}, x^{k} \rangle]^{-1}y^{k}$
        \State Initialize $\nu_{R}^{k}:$ Equation \eqref{KSR}
    \Else
        \State Find the most compatible rule: $i = \arg \max_{i} \{c_{i}^{k}\}$
        \State Update the rule center: $\upsilon_{i}^{k} = \upsilon_{i}^{k - 1} + \alpha(c_{i}^{k})^{(1 - a_{i}^{k})} (x^{k} - \upsilon_{i}^{k - 1})$
        \State Compute $g, z$ and $r$: $g^{k} = [\kappa \langle d_{i1}^{k}, x^{k} \rangle, \ldots , \kappa \langle d_{in i}^{k}, x^{k} \rangle]^{T}, z^{k} = Q_{i}^{k - 1} g ^{k},$ $r^{k} = \lambda + \kappa \langle x^{k}, x^{} \rangle - (z^{k})^{T} g^{k}$
               
            \If {$\min_{(\forall d_{ij} \in \mathcal{D}_{i}^{k})} \| x^{k} - d_{ij}^{k} \| \geq 0.1 \nu_{ij}^{k}$}
                \State Include $x^{k}$ into the dictionary: $\mathcal{D}_{i}^{k} = \mathcal{D}_{i}^{k} \cup x^{k}$
                \State Compute $Q_{i}^{k}, P_{i}^{k}, \theta_{i}^{k}, \nu_{i}^{k}$: Equations \eqref{UP}, \eqref{P}, \eqref{CP}, \eqref{KSU}
            \Else
                \State Compute $Q_{i}^{k}, P_{i}^{k}, q_{i}^{k}$ and $\theta_{i}^{k}$: Equations \eqref{Q}, \eqref{P eq}, \eqref{q}, \eqref{theta}
            \EndIf
    \EndIf
    
    \For {$i = 1, 2, \ldots, R$}
        \If {$U_{i}^{k} < \epsilon$}
            \State Remove underused rules: Remove $(i)$
            \State $Excluded\_Rule == true$
        \EndIf
    \EndFor
    
    \State Compute the output: $\hat{y} = \Sigma_{j = 1}^{n_i} \theta_{ij} \kappa \langle d_{ij}^{k}, x^{k} \rangle \vert i = \arg \max_{i} \{c_{i}^{k}\} $
\EndFor
\end{algorithmic}
\end{algorithm}

\begin{algorithm}
\caption{Generation of fuzzy sets algorithm} \label{fuzzycreator}
\begin{algorithmic}[1]
\renewcommand{\algorithmicrequire}{\textbf{Input:}}
\renewcommand{\algorithmicensure}{\textbf{Output:}}

\Require $x^{k}, \upsilon_{i}^{k}, A, B, T$
\Ensure  $\hat{\mu}_{A}, \hat{\mu}_{B} \ \textbf{or} \ \hat{\mu}_{\tilde{A}}, \hat{\mu}_{\tilde{B}}$

\State \textbf{Initialisation}: $\mathcal{I} = Linear, N = False, \mathcal{T}^{1} = 0$ 

\For {$k = 2, 3, \ldots, n$}
    \For {$i = 1, 2, \ldots, R$}
        \For {$p = 1, 2, \ldots, m$}
            \If {$T == \textit{Type-2}$}
                \State Split the data in two subsets:
                \If {$m \ \% \ 2 \ \neq 0$}
                    \State Create the subsets: Equation \eqref{subset}
                \Else
                    \State Create the subsets with repeated value: Equation \eqref{subset-odd}
                \EndIf
                \State Calculate membership values for each subset: Equation \eqref{MFT2}
            \EndIf
            \If {$A  \ \textbf{or} \ B == Gaussian$}
                \State Create the endpoints of the MFs: Equation \eqref{Y min max}
                \If {$(x_{p})^{k} \ \textbf{or} \ (\upsilon_{p})_{i}^{k} \notin \mathcal{P}^{k}$}
                    \State $\mu_A (x^{k}) =  0 \ \textbf{or} \ \mu_B (\upsilon_{i}^{k}) = 0$
                \Else
                    \State Calculate the membership values: Equation \eqref{MF}
                \EndIf
            \EndIf
            \If {$A  \ \textbf{or} \ B == Discrete$}
                \State Calculate the membership values: Equation \eqref{MFD}
            \EndIf
            \If {$A  \ \textbf{or} \ B == Polling$}
                \State Calculate the membership values: Equation \eqref{MFP}
            \EndIf
            \If {$A  \ \textbf{or} \ B == Interval \ Polling$}
                \For {$\mathcal{I} = 0, 1, \ldots, m$}
                    \State Calculate the membership values: Equation \eqref{MFI}
                \EndFor
            \EndIf
        \EndFor
    \EndFor
    \State Compute the outputs: 
    \If {$T == \textit{Type-1}$} 
        \State $\hat{\mu}_{A} = \Sigma_{j = 1}^{n_i} \langle \mu_{A}(x^{k})\rangle$ \ ; \ $\hat{\mu}_{B} = \Sigma_{j = 1}^{n_i} \langle \mu_{B}(\upsilon_{i}^{k})\rangle$
    \EndIf
    \If {$T == \textit{Type-2}$} 
        \State $\hat{\mu}_{\tilde{A}} = \Sigma_{j = 1}^{n_i} \langle \mu_{\tilde{A}}({}_{1}x^{k}, {}_{2}x^{k})\rangle$ \ ; \ $\hat{\mu}_{\tilde{B}} = \Sigma_{j = 1}^{n_i} \langle \mu_{\tilde{B}}({}_{1}\upsilon_{i}^{k}, {}_{2}\upsilon_{i}^{k})\rangle$
    \EndIf
\EndFor
\end{algorithmic}
\end{algorithm}

\begin{table}[htbp]
\caption{Description of the parameters}
\begin{center}
\begin{tabular}{ p{1.35cm} p{6.15cm} }
\hline
    Parameter & Description \\
\hline
    $x$ & Input vector\\
    $y$ &  Desired output\\
    $\alpha$ &  Learning rate\\
    $\beta$ &   Parameter that controls the growth
rate of $a$\\
    $\lambda$ & Normalized activation level\\
    $\tau$ & Activation level \\
    $\omega$ & Weight vector\\
    $\sigma$ & Spread of
the antecedent part\\
    $\epsilon$ & Threshold of the utility measure\\
    $A$ & Fuzzy set of the input vector\\
    $B$ & Fuzzy set of the cluster center\\
    $M$ & Measure adopted\\
    $R$ & Number of rules\\
    $\upsilon$ & Center of the rule\\
    $\mathcal{D}$ & Local dictionary\\
    $\nu$ & Kernel size of the dictionary\\
    $P$ & Covariance matrix\\
    $\theta$ & Consequent parameters\\
    $\kappa$ & Gaussian-Kernel function\\
    $a$ & Arousal index\\
    $c$ & Compatibility measure\\
    $g$ & Parameter that calculates the $\theta$ \\
    $z$ & Parameter that calculates the $\theta$ \\
    $r$ & Parameter that calculates the $\theta$ \\
    $Q$ & Parameter that calculates the $\theta$ \\
    $U$ & Utility measure\\
    $T$ & Parameter that defines the type of fuzzy set\\
    $\mu_A$ & Membership function of the input vector\\
    $\mu_B$ & Membership function of the cluster center\\
    $\mathcal{I}$ & Interval of the actual value of the fuzzy set \\
    $N$ & Normalization of the fuzzy set\\
    $\mathcal{T}$ & Number of total intervals of the actual value\\    
\hline 
\end{tabular}
\end{center}
\label{param-algo}
\end{table}

%% Experimental Results
\section{Experimental Results}\label{Results}

The root-mean-square error (RMSE), non-dimensional index error (NDEI), and mean absolute error (MAE) are implemented to evaluate the model's performance for the Mackey-Glass chaotic time series, using Equations \eqref{RMSE}, \eqref{NDEI} and \eqref{MAE}, respectively.

\begin{equation} \label{RMSE}
    RMSE = \sqrt{ \dfrac{1}{T} \sum_{k = 1}^{T} (y^{k} - \hat{y}^{k})^{2}}
\end{equation}

\begin{equation} \label{NDEI}
    NDEI = \dfrac{RMSE}{std([y^{1}, \ldots, y^{T}])}
\end{equation}

\begin{equation} \label{MAE}
    MAE = \dfrac{1}{T} \sum_{k = 1}^{T} \vert y^{k} - \hat{y}^{k} \vert
\end{equation}

\noindent where $T$ is the sample size, $y^{k}$ is the $k$th actual value, $\hat{y}^{k}$ is the $k$th predicted value and $std()$ is the standard deviation function.

As RMSE and MAE do not calculate the effects of existing errors within different periods and do not consider the direction of predictions (\citealt{ramezanian2019integrated}), the error of coefficient of multiple determinations ($ER^2$), NDEI, and the mean absolute percentage error (MAPE) are implemented to evaluate the model's performance for the TAIEX main stock index. The $ER^2$ and the MAPE are calculated according to Equations \eqref{R2} and \eqref{MAPE}, respectively.

\begin{equation} \label{R2}
    ER^2 = \dfrac{\sum_{k = 1}^{T} (y^{k} - \hat{y}^{k})^{2}}{\sum_{k = 1}^{T} (y^{k} - \bar{y}^{k})^{2}}
\end{equation}

\begin{equation} \label{MAPE}
    MAPE = \dfrac{1}{T} \sum_{k = 1}^{T} \dfrac{\vert y^{k} - \hat{y}^{k} \vert}{\max(\epsilon, \vert \hat{y}^{k} \vert)}
\end{equation}

\noindent where $\bar{y}^{k}$ is the mean value of all measured data points, and $\epsilon$ is an arbitrarily small yet strictly positive number to avoid undefined results when $y$ is zero. The $ER^2$ is based on the traditional coefficient of determination ($R^2$). These equations are used because they are more appropriate for stock exchange predictions (\citealt{boyacioglu2010adaptive, chaigusin2008use, rimcharoen2005prediction}).

Another way to measure the computational complexity is using time (\citealt{oliveto2011runtime}). The execution time is estimated in seconds, computing the mean runtime and the standard deviation of twenty simulations. Furthermore, the number of final rules is presented in each model. The results of ePL-KRLS-FSM+ are compared with other evolving fuzzy modeling approaches, such as ePL-KRLS-FSM (\citealt{de2022kernel, marques2022kernel}), ePL-KRLS (\citealt{alves2020enhanced}), ePL-KRLS-DISCO (\citealt{alves2021novel}) and eMG (\citealt{lemos2010multivariable}) models, and with traditional forecasting models such as  Decision Trees (DT) (\citealt{breiman1984classification}), Multi-Layer Perceptron (MLP) (\citealt{rumelhart1986learning}) and Support Vector Machine (SVM) (\citealt{smola2004tutorial}). All codes were executed using Python 3.9 in a PC device with Intel Core Celeron CPU 3215U, 1.70 GHz, and 8 GB RAM.

Normalization is performed to avoid the varying scales of the forecasting model (\citealt{hewamalage2021global}), and is based on the maximum and minimum values, as follows:

\begin{equation}
    N_{norm}(x) = \dfrac{x - \min(x)}{\max(x) - \min(x)}
\end{equation}

The hyperparameters were previously defined by \citet{de2022kernel, marques2022kernel} to produce the lowest errors. For the ePL-KRLS models (ePL-KRLS, ePL-KRLS-DISCO, ePL-KRLS-FSM and ePL-KRLS-FSM+) the following general guidelines are suggested: $\alpha \leq 0.1$, $\beta \in [0, 1]$, $\lambda \in [0, 1]$, $\sigma \in [0.2, 0.5]$, $\omega = 1$, and $\epsilon \in [0.03, 0.05]$. The parameters of the fuzzy sets (normalization, universe of discourse, and the number of discretizations) are set by default according to \citet{mcculloch2017fuzzycreator}. All the FSs are generated as Gaussian Functions.

%% Mackey-Glass chaotic time series
\subsection{Mackey-Glass chaotic time series}

The model is evaluated using the Mackey-Glass chaotic time series (\citealt{mackey1977oscillation}), according to the following differential equation:

\begin{equation} \label{MackeyGlass}
    \dfrac{dx(t)}{dt} = \varphi \dfrac{ x (t - \vartheta)}{1 + x^{\varrho} (t - \vartheta)} - \varsigma x(t - 1)
\end{equation}

\noindent where $t$ is the time, $\vartheta$ is the delay, and $\varphi = 0.2$, $\varrho = 10$, and $\varsigma = 0.1$ are parameters that determine the periodicity and the chaos induced into the resulting series (\citealt{hewamalage2021global}). To induce a chaotic behavior, we adopt $\vartheta > 17$ and $\varrho \approx 10$ (\citealt{MackeyGlass:CVUT}).

The objective is predict $x^{k + 85}$ using as input vector $[x^{k}, x^{k + 6}, x^{k + 12}, x^{k + 18}]$ for any $k$ value. In the training phase, 3000 data samples were used for $k \in [201, 3200]$, and 500 data samples were selected to test the model for $k \in [5001, 5500]$ (\citealt{de2022kernel}). To make the series chaotic it is necessary to change $\vartheta$, so the values adopted are $\vartheta = [17, 500, 1000]$. Table \ref{hyper-mackey} presents the hyperparameters for the Mackey-Glass time series.

\begin{table}[htbp]
\caption{Hyperparameters of the models for the Mackey-Glass time series}
\begin{center}
\begin{tabular}{ p{2.736cm} p{5cm} }
\hline
    Model & Hyperparameters \\
\hline
    DT (\citealt{breiman1984classification}) & Two of maximum depth of the tree\\
    MLP (\citealt{rumelhart1986learning}) & A hidden layer with three neurons\\
    SVM (\citealt{smola2004tutorial}) & $C = 1$, $\gamma = 1$ \\
    eMG (\citealt{lemos2010multivariable}) & $\alpha = 0.1$, $\lambda = 0,1$, $w = 10$, $\sigma = 0.05$, $\omega = 10^{4}$ \\
    ePL-KRLS (\citealt{maciel2017evolving}) & $\alpha = 0.1$, $\beta = 0.09$, $r = 0.5$, $\gamma = 0.91$\\
    ePL-KRLS-DISCO (\citealt{alves2021novel}) & $\alpha = 0.001$, $\beta = 0.06$, $\lambda = 10^{-7}$, $\sigma = 0.3$,  $\omega = 1$, $\epsilon = 0.05$\\
    ePL-KRLS-FSM & $\alpha = 0.001$, $\beta = 0.06$, $\lambda = 10^{-7}$, $\sigma = 0.3$,  $\omega = 1$, $\epsilon = 0.05$, $A = Gaussian$, $B = Gaussian$\\
\hline 
\end{tabular}
\end{center}
\label{hyper-mackey}
\end{table}

Table \ref{17} shows the model's results with $\vartheta = 17$. The ePL-KRLS-FSM\_giq model obtained the lowest errors, and the ePL-KRLS-FSM\_p3, ePL-KRLS-T2FSM\_zl, ePL-KRLS-T2FSM\_ma and ePL-KRLS-T2FSM\_hy presented one final rule in the simulations. The eMG obtained the highest number of final rules and the ePL-KRLS presented the highest errors.

\begin{table*}[htbp]
\caption{Simulations results of Mackey-Glass time series with $\vartheta = 17$}
\begin{center}
\resizebox{\linewidth}{!}{%
\begin{tabular}{ c c c c c }
\hline
    Measure & RMSE & NDEI & MAE & Rules \\
\hline
    DT (\citealt{breiman1984classification}) & 0.0745571 & 0.3173367 & 0.0609805 & 4 \\
    MLP (\citealt{rumelhart1986learning}) & 0.0665699 & 0.2833410 & 0.0553101 & - \\
    SVM (\citealt{smola2004tutorial}) & 0.0609649 & 0.2594842 & 0.0541051 & - \\
    eMG (\citealt{lemos2010multivariable}) & 0.0201885 & 0.0859283 & 0.0135714 & 42 \\
    ePL-KRLS (\citealt{maciel2017evolving}) & 0.2861642 & 1.2179974 & 0.2586356 & 3 \\
    ePL-KRLS-DISCO (\citealt{alves2021novel}) & 0.0015681 & 0.0066745 & 0.0005112 & 13 \\
    ePL-KRLS-FSM\_p1 (\citealt{de2022kernel}) & 0.0036433 & 0.0155069 & 0.0012803 & 15 \\
    ePL-KRLS-FSM\_p3 (\citealt{de2022kernel}) & 0.0005173 & 0.0022016 & 0.0003752 & \textbf{1} \\
    ePL-KRLS-FSM\_zw (\citealt{de2022kernel}) & 0.0032723 & 0.0139280 & 0.0010409 & 14 \\
    ePL-KRLS-FSM\_r2 (\citealt{marques2022kernel}) & 0.0004849 & 0.0020637 & 0.0003467 & 15 \\
    ePL-KRLS-FSM\_cr (\citealt{marques2022kernel}) & 0.0016561 & 0.0070487 & 0.0007574 & 13 \\
    ePL-KRLS-FSM\_giq (\citealt{marques2022kernel}) & \textbf{0.0004625} & \textbf{0.0019687} & \textbf{0.0003226} & 14 \\
    ePL-KRLS-T1FSM\_mc & 0.0014594 & 0.0062137 & 0.0010809 & 2 \\
    ePL-KRLS-T2FSM\_zl & 0.0005451 & 0.0023209 & 0.0003896 & \textbf{1} \\
    ePL-KRLS-T2FSM\_jg2 & 0.0022384 & 0.0095308 & 0.0009578 & 14 \\
    ePL-KRLS-T2FSM\_zc & 0.0026400 & 0.0112406 & 0.0010622 & 14 \\
    ePL-KRLS-T2FSM\_hc & 0.0013845 & 0.0058949 & 0.0007528 & 14 \\
    ePL-KRLS-T2FSM\_yl & 0.0008437 & 0.0035922 & 0.0005527 & 14 \\
    ePL-KRLS-T2FSM\_ma & 0.0005451 & 0.0023209 & 0.0003896 & \textbf{1} \\
    ePL-KRLS-T2FSM\_hy & 0.0005451 & 0.0023209 & 0.0003896 & \textbf{1} \\
\hline 
\end{tabular}}
\end{center}
\label{17}
\end{table*}

Table \ref{500} presents the results of simulations for $\vartheta = 500$. The results show that \textit{zl}, \textit{ma} and \textit{hy} Fuzzy Sets Measures (FSM) performed better in terms of RMSE, NDEI, and the number of final rules, achieving the same results. The ePL-KRLS-FSM\_p1 obtained the best MAE error. The eMG model obtained the highest number of final rules again and the ePL-KRLS presented the highest errors once again.

\begin{table*}[htbp]
\caption{Simulations results of Mackey-Glass time series with $\vartheta = 500$}
\begin{center}
\resizebox{\linewidth}{!}{%
\begin{tabular}{ c c c c c }
\hline
    Measure & RMSE & NDEI & MAE & Rules \\
\hline
    DT (\citealt{breiman1984classification}) & 0.0913939 & 0.2869591 & 0.0765731 & 4 \\
    MLP (\citealt{rumelhart1986learning}) & 0.1417790 & 0.4451588 & 0.1265467 & - \\
    SVM (\citealt{smola2004tutorial}) & 0.0549357 & 0.1724874 & 0.0467751 & - \\
    eMG (\citealt{lemos2010multivariable}) & 0.0249023 & 0.0781883 & 0.0184309 & 62 \\
    ePL-KRLS (\citealt{maciel2017evolving}) & 0.2779180 & 0.8726090 & 0.2533209 & 4 \\
    ePL-KRLS-DISCO (\citealt{alves2021novel}) & 0.0046359 & 0.0145559 & 0.0025172 & 12 \\
    ePL-KRLS-FSM\_p1 (\citealt{de2022kernel}) & 0.0043928 & 0.0137927 & \textbf{0.0022323} & 13 \\
    ePL-KRLS-FSM\_p3 (\citealt{de2022kernel}) & 0.0032873 & 0.0103216 & 0.0023416 & \textbf{1} \\
    ePL-KRLS-FSM\_zw (\citealt{de2022kernel}) & 0.0042624 & 0.0133831 & 0.0029777 & 14 \\
    ePL-KRLS-FSM\_r2 (\citealt{marques2022kernel}) & 0.0037423 & 0.0117502 & 0.0025763 & 11 \\
    ePL-KRLS-FSM\_cr (\citealt{marques2022kernel}) & 0.0039184 & 0.0123029 & 0.0022737 & 13 \\
    ePL-KRLS-FSM\_giq (\citealt{marques2022kernel}) & 0.0039139 & 0.0122888 & 0.0028218 & 10 \\
    ePL-KRLS-T1FSM\_mc & 0.0043372 & 0.0136181 & 0.0034042 & 5 \\
    ePL-KRLS-T2FSM\_zl & \textbf{0.0032835} & \textbf{0.0103095} & 0.0023375 & \textbf{1} \\
    ePL-KRLS-T2FSM\_jg2 & 0.0195869 & 0.0614992 & 0.0043759 & 13 \\
    ePL-KRLS-T2FSM\_zc & 0.0450591 & 0.1414768 & 0.0049643 & 13 \\
    ePL-KRLS-T2FSM\_hc & 0.0051900 & 0.0162956 & 0.0031502 & 13 \\
    ePL-KRLS-T2FSM\_yl & 0.0059070 & 0.0185470 & 0.0036799 & 12 \\
    ePL-KRLS-T2FSM\_ma & \textbf{0.0032835} & \textbf{0.0103095} & 0.0023375 & \textbf{1} \\
    ePL-KRLS-T2FSM\_hy & \textbf{0.0032835} & \textbf{0.0103095} & 0.0023375 & \textbf{1} \\
\hline 
\end{tabular}}
\label{500}
\end{center}
\end{table*}

\begin{figure}[htbp]
\centerline{\includegraphics[scale=0.2, trim = 0cm 0cm 2.3cm 3.3cm, clip]{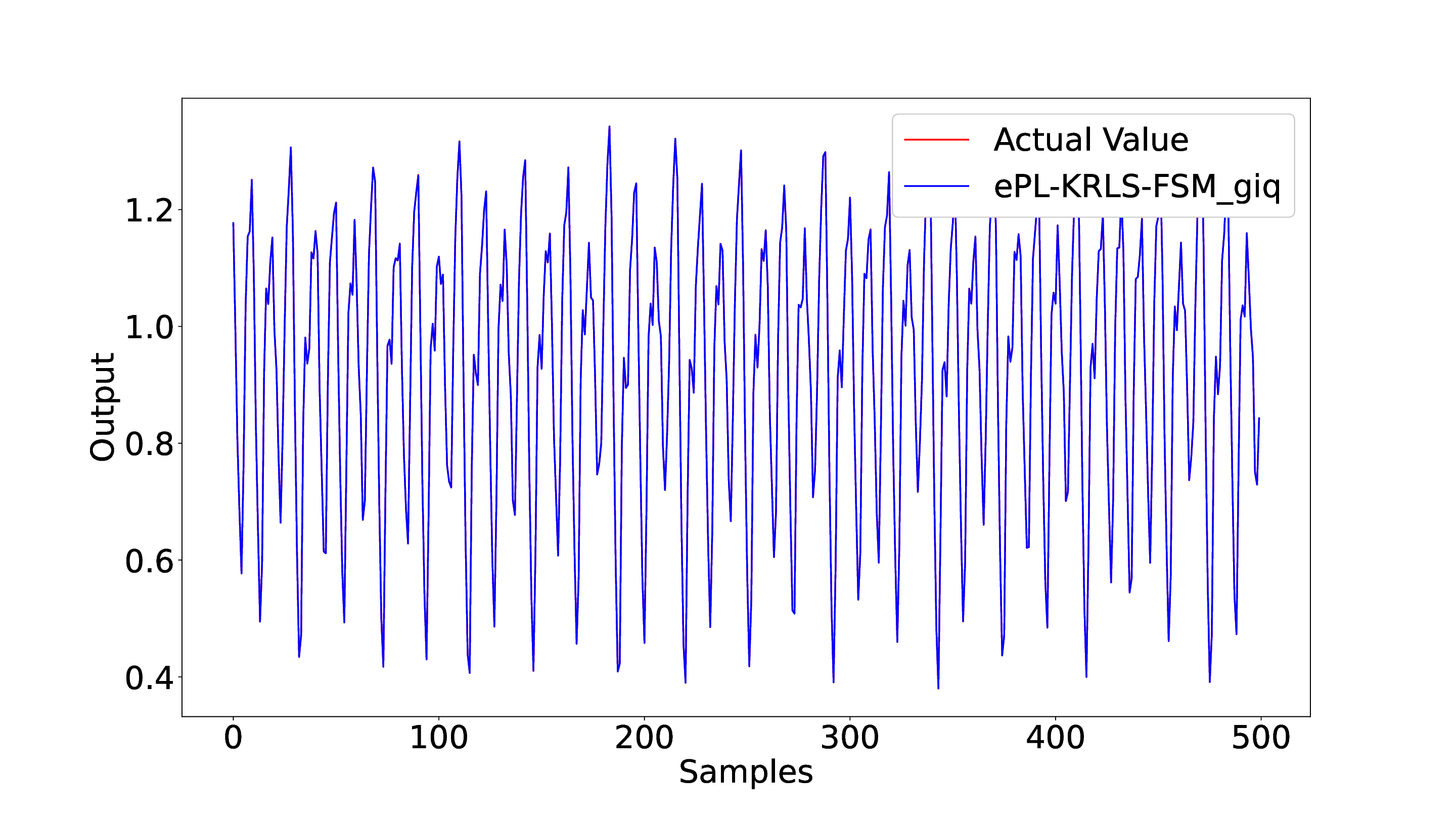}}
\caption{Predictions of ePL-KRLS-FSM with $\vartheta = 17$.}
\label{pappis3 17}
\end{figure}

\begin{figure}[htbp]
\centerline{\includegraphics[scale=0.2, trim = 0cm 0cm 2.3cm 3.3cm, clip]{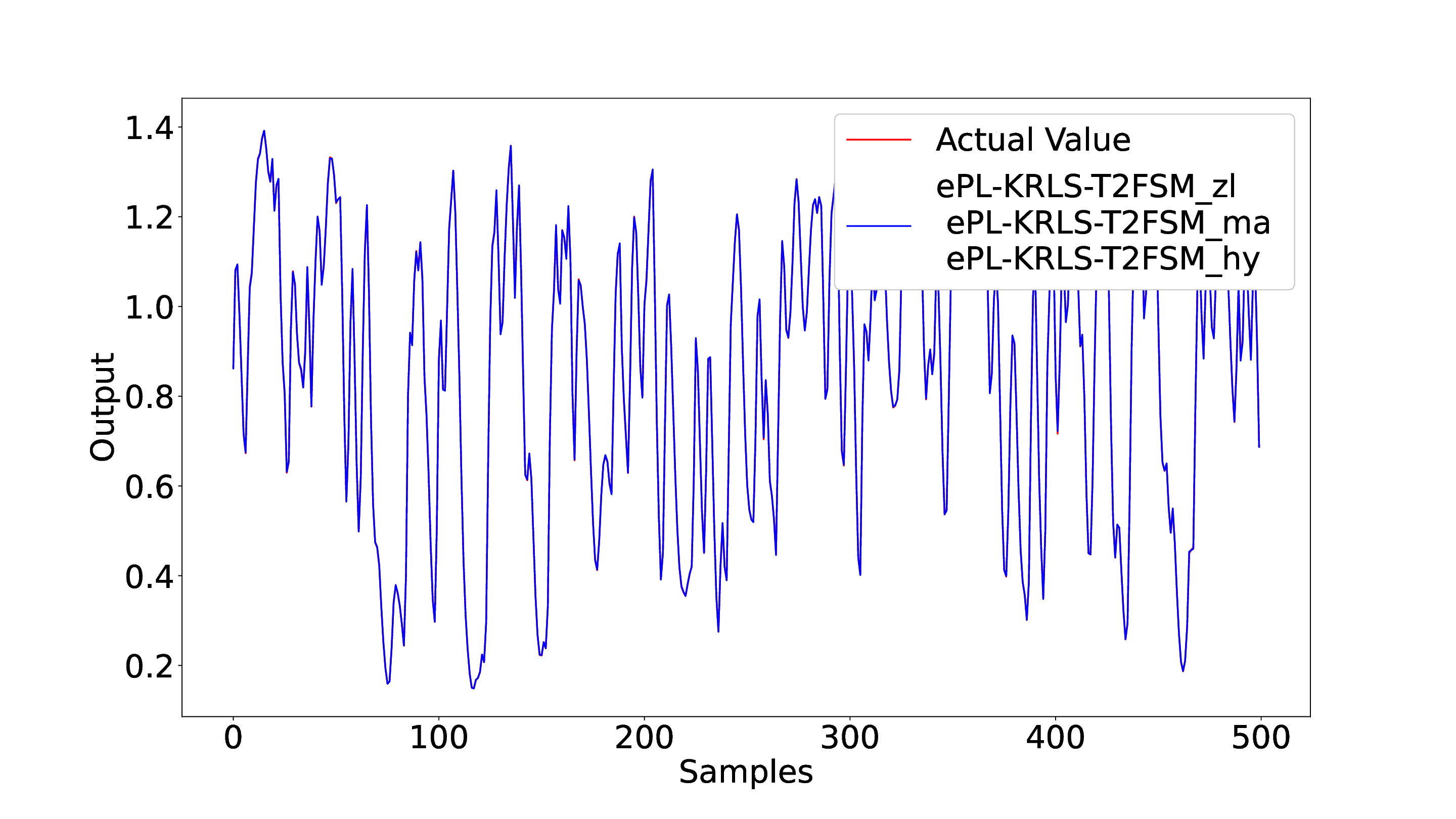}}
\caption{Predictions of ePL-KRLS-FSM+ with $\vartheta = 500$.}
\label{pappis3 500}
\end{figure}

\begin{figure}[htbp]
\centerline{\includegraphics[scale=0.2, trim = 0cm 0cm 2.3cm 3.3cm, clip]{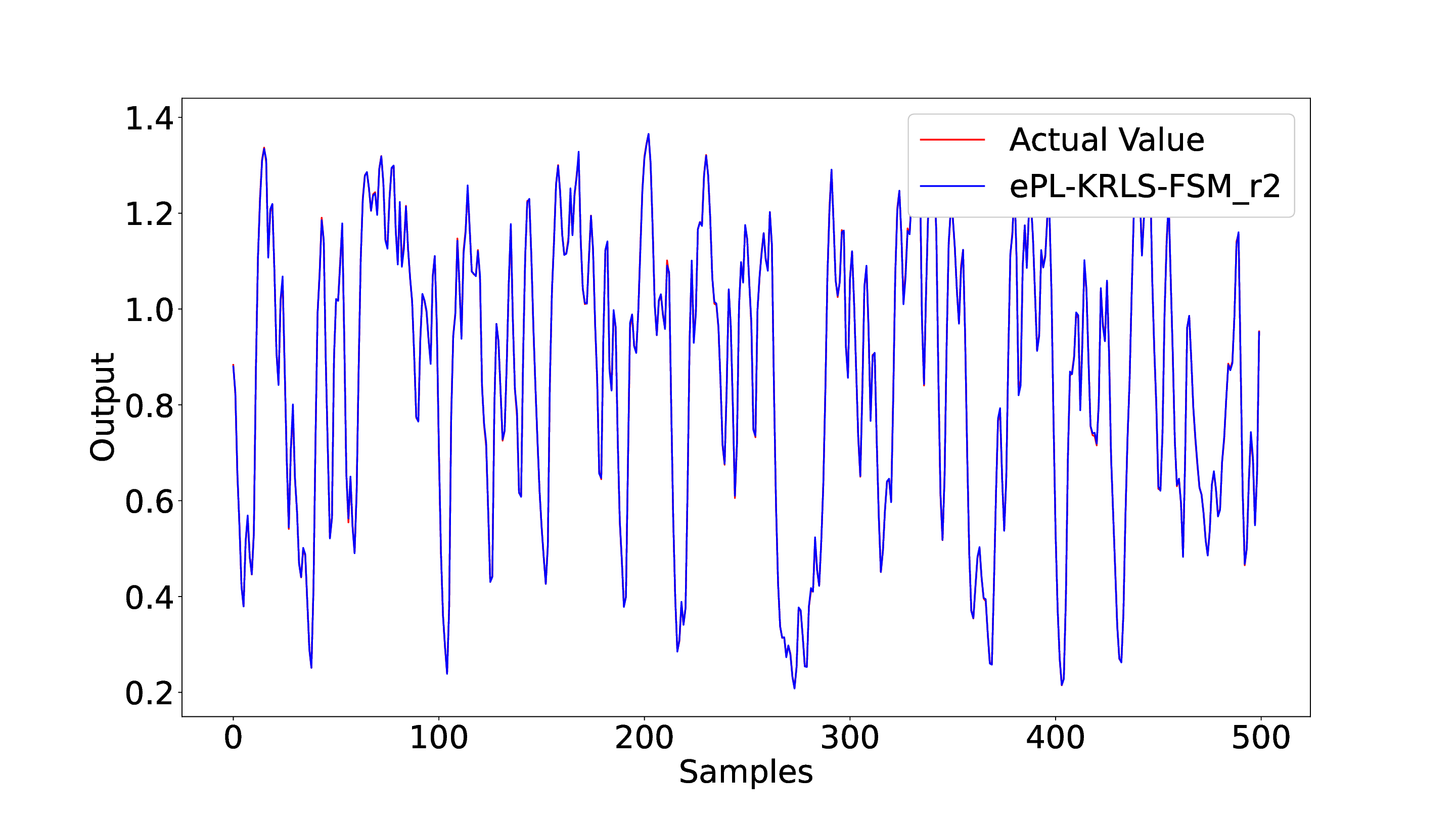}}
\caption{Predictions of ePL-KRLS-FSM with $\vartheta = 1000$.}
\label{pappis1 1000}
\end{figure}

Table \ref{1000} shows the simulations' results for $\vartheta = 1000$. ePL-KRLS-FSM\_r2 obtained the lowest RMSE and NDEI among all models, and the ePL-KRLS-FSM\_p1 obtained the lowest MAE error. The eMG model obtained the highest number of final rules and the ePL-KRLS presented one more time the worst results, and \textit{pappis3}, \textit{zeng\_li}, \textit{mohamed\_abdaala} and \textit{hung\_yang} achieved the lowest number of final rules again. Figures~\ref{pappis3 17}, \ref{pappis3 500}, and \ref{pappis1 1000} show the plots of the best model for all $\vartheta$ values.

\begin{table*}[htbp]
\caption{Simulations results of Mackey-Glass time series with $\vartheta = 1000$}
\begin{center}
\resizebox{\linewidth}{!}{%
\begin{tabular}{ c c c c c }
\hline
    Model & RMSE & NDEI & MAE & Rules \\
\hline
    DT (\citealt{breiman1984classification}) & 0.0841051 & 0.2813752 & 0.0708099 & 4 \\
    MLP (\citealt{rumelhart1986learning}) & 0.0566753 & 0.1896084 & 0.0476612 & - \\
    SVM (\citealt{smola2004tutorial}) & 0.0627334 & 0.2098757 & 0.0515686 & - \\
    eMG (\citealt{lemos2010multivariable}) & 0.0246279 & 0.0823930 & 0.0189387 & 83 \\
    ePL-KRLS (\citealt{maciel2017evolving}) & 0.2868170 & 0.9595521 & 0.2609193 & 3 \\
    ePL-KRLS-DISCO (\citealt{alves2021novel}) & 0.0062104 & 0.0207771 & 0.0033879 & 13 \\
    ePL-KRLS-FSM\_p1 (\citealt{de2022kernel}) & 0.0042029 & 0.0140608 & \textbf{0.0024044} & 12 \\
    ePL-KRLS-FSM\_p3 (\citealt{de2022kernel}) & 0.0047771 & 0.0159820 & 0.0034942 & \textbf{1} \\
    ePL-KRLS-FSM\_zw (\citealt{de2022kernel}) & 0.0050182 & 0.0167884 & 0.0034500 & 14 \\
    ePL-KRLS-FSM\_r2 (\citealt{marques2022kernel}) & \textbf{0.0040003} & \textbf{0.0133832} & 0.0028836 & 11 \\
    ePL-KRLS-FSM\_cr (\citealt{marques2022kernel}) & 0.0113861 & 0.0380925 & 0.0046725 & 12 \\
    ePL-KRLS-FSM\_giq (\citealt{marques2022kernel}) & 0.0083742 & 0.0280161 & 0.0032480 & 11 \\
    ePL-KRLS-T1FSM\_mc & 0.0050564 & 0.0169162 & 0.0036790 & 5 \\
    ePL-KRLS-T2FSM\_zl & 0.0047661 & 0.0159451 & 0.0034617 & \textbf{1} \\
    ePL-KRLS-T2FSM\_jg2 & 0.0115559 & 0.0386606 & 0.0046225 & 13 \\
    ePL-KRLS-T2FSM\_zc & 0.0059082 & 0.0197660 & 0.0039242 & 12 \\
    ePL-KRLS-T2FSM\_hc & 0.0129500 & 0.0433244 & 0.0045102 & 12 \\
    ePL-KRLS-T2FSM\_yl & 0.0071534 & 0.0239319 & 0.0046342 & 13 \\
    ePL-KRLS-T2FSM\_ma & 0.0047661 & 0.0159451 & 0.0034617 & \textbf{1} \\
    ePL-KRLS-T2FSM\_hy & 0.0047661 & 0.0159451 & 0.0034617 & \textbf{1} \\
\hline 
\end{tabular}}
\label{1000}
\end{center}
\end{table*}

The results demonstrate that ePL-KRLS-FSM+ is an effective model for predicting chaotic time series. The ePL-KRLS-FSM\_p3, ePL-KRLS-T2FSM\_zl, ePL-KRLS-T2FSM\_ma and ePL-KRSL-FSM+\_hy presented only one final rule in all simulations, indicating a low computational cost (\citealt{alves2020enhanced}). The \textit{zl}, \textit{ma} and \textit{hy} T2FSMs achieved accurate outputs for different $\vartheta$ values, obtaining the lowest errors for $\vartheta = 500$ among the models. The ePL-KRLS-FSM\_giq obtained the best results for $\vartheta = 17$, and the ePL-KRLS-FSM\_r2 for $\vartheta = 1000$. The ePL-KRLS-T2FSM\_zc had a significant improvement in the results for $\vartheta = 1000$. And the \textit{zw}, \textit{mc}, \textit{yl} FSMs showed satisfactory error results.

%% Financial
\subsection{Stock Market}

The Taiwan Stock Exchange Capitalization Weighted Stock Index (TAIEX) (\citealt{TAIEX}) is also applied to evaluate the models' performance. The datasets are Close Values of 1-day interval trading exchange from 2001 to 2015 (\citealt{marques2022kernel}). The evolving fuzzy systems obtain the following formulation to evaluate the time series forecasting of the main stock index (\citealt{maciel2017evolving}):

\begin{equation} \label{SME}
    y_{t} = f(y_{t-1}, y_{t-2}, \ldots, y_{t-p})
\end{equation}

\noindent where $p=4$ is the number of lags. The aim is to predict $y^{k+4}$ using as input vector $[y^{k}, y^{k+2}, y^{k+3}]$ for any $k$ value. 3200 data samples were used for $k \in [1, 3200]$ to train the model, and 300 data samples were used in testing phase for $k \in [3201, 3500]$. Table \ref{hyper-TAIEX} presents the hyperparameters for the TAIEX Stock Exchange time series.

\begin{table}[htbp]
\caption{Hyperparameters of the model's for the TAIEX Stock Exchange time series}
\begin{center}
\begin{tabular}{ p{2.736cm} p{5cm} }
\hline
    Model & Hyperparameters \\
\hline
    DT (\citealt{breiman1984classification}) & Eight of maximum depth of the tree\\
    MLP (\citealt{rumelhart1986learning}) & A hidden layer with twenty neurons\\
    SVM (\citealt{smola2004tutorial}) & $C = 1$, $\gamma = 1$ \\
    eMG (\citealt{lemos2010multivariable}) & $\alpha = 0.1$, $\lambda = 0.01$, $w = 10$, $\sigma = 10^{-2}$, $\omega = 10^{4}$ \\
    ePL-KRLS (\citealt{maciel2017evolving}) & $\alpha = 0.1$, $\beta = 0.24$, $r = 0.5$, $\gamma = 0.76$\\
    ePL-KRLS-DISCO (\citealt{alves2021novel}) & $\alpha = 0.01$, $\beta = 0.1$, $\lambda = 10^{-3}$, $\sigma = 0.5$,  $\omega = 1$, $\epsilon = 0.05$\\
    ePL-KRLS-FSM & $\alpha = 0.01$, $\beta = 0.1$, $\lambda = 10^{-3}$, $\sigma = 0.5$,  $\omega = 1$, $\epsilon = 0.05$, $A = Gaussian$, $B = Gaussian$\\
\hline 
\end{tabular}
\end{center}
\label{hyper-TAIEX}
\end{table}

\begin{figure}[htbp]
\centerline{\includegraphics[scale=0.2, trim = 0cm 0cm 2.3cm 3.3cm, clip]{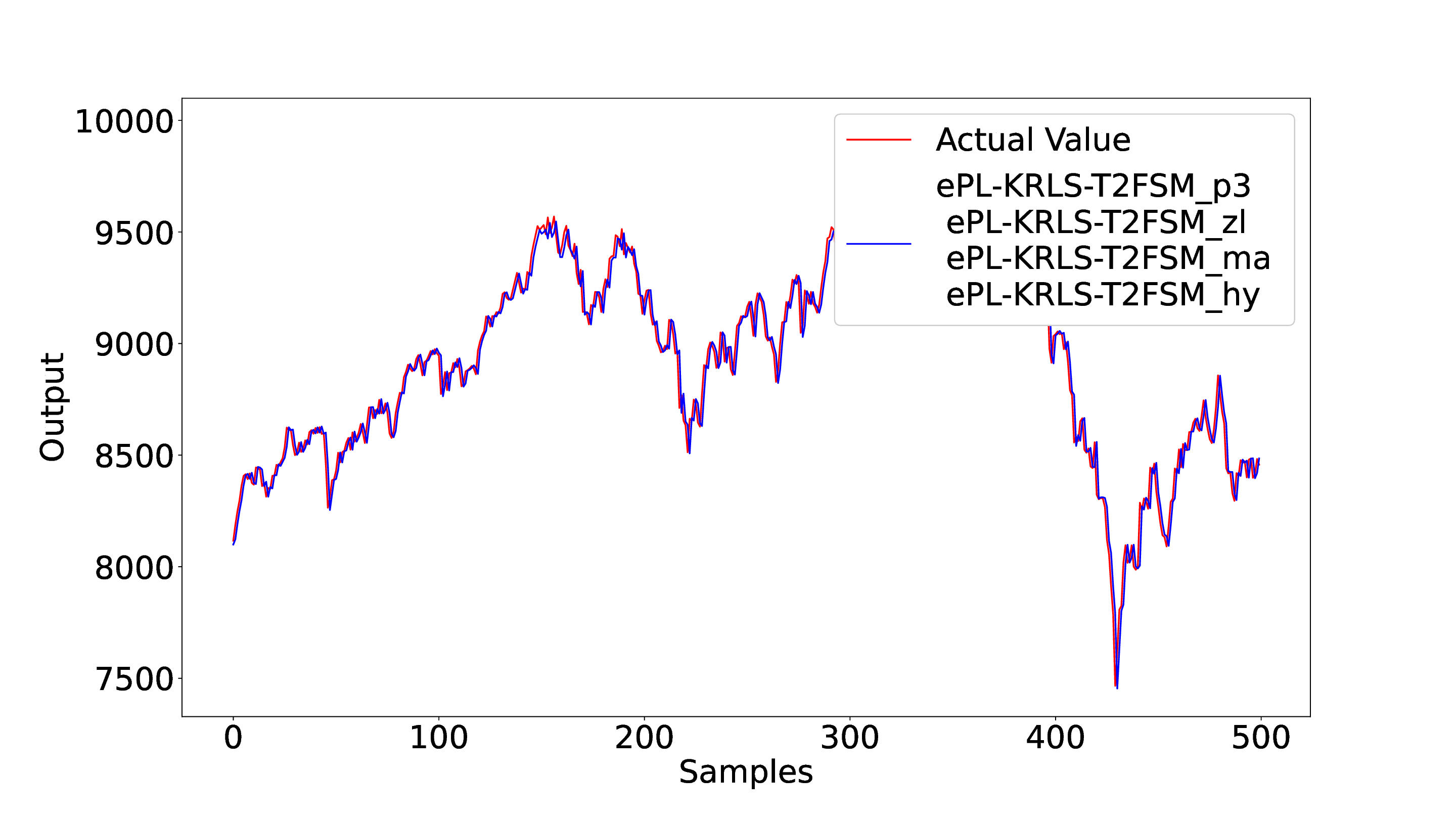}}
\caption{Predictions of ePL-KRLS-FSM+ with the main stock index.}
\label{TX}
\end{figure}

Table \ref{TAIEX} shows the simulations' results for the stock market time series. The \textit{pappis3}, \textit{zeng\_li}, \textit{mohamed\_abdaala} and \textit{hung\_yang} models achieved the lowest errors among the models, and the lowest number of final rules along with eMG and the ePL-KRLS models. Figure \ref{TX} shows the plot of the best model for the main stock index.

\begin{table*}[htbp]
\caption{Simulations results of TAIEX time series}
\begin{center}
\resizebox{\linewidth}{!}{%
\begin{tabular}{ c c c c c }
\hline
    Measure & $ER^2$ & NDEI & MAPE & Rules \\
\hline
    DT (\citealt{breiman1984classification}) & 0.0801730 & 0.2826760 & 0.0124152 & 224 \\
    MLP (\citealt{rumelhart1986learning}) & 0.0586197 & 0.2417111 & 0.0114034 & - \\
    SVM (\citealt{smola2004tutorial}) & 0.2900036 & 0.5376215 & 0.0241796 & - \\
    eMG (\citealt{lemos2010multivariable}) & 0.0943168 & 0.3065981 & 0.0145756 & \textbf{1} \\
    ePL-KRLS (\citealt{maciel2017evolving}) & 0.0549435 & 0.2340094 & 0.0109190 & \textbf{1} \\
    ePL-KRLS-DISCO (\citealt{alves2021novel}) & 0.0342947 & 0.1848795 & 0.0081511 & 14 \\
    ePL-KRLS-FSM\_p1 (\citealt{de2022kernel}) & 0.0476401 & 0.2179021 & 0.0097729 & 14 \\
    ePL-KRLS-FSM\_p3 (\citealt{de2022kernel}) & \textbf{0.0258646} & \textbf{0.1605565} & \textbf{0.0070144} & \textbf{1} \\
    ePL-KRLS-FSM\_zw (\citealt{de2022kernel}) & 0.0581158 & 0.2406702 & 0.0109078 & 17 \\
    ePL-KRLS-FSM\_r2 (\citealt{marques2022kernel}) & 0.0285373 & 0.1686480 & 0.0073180 & 9 \\
    ePL-KRLS-FSM\_cr (\citealt{marques2022kernel}) & 0.0283965 & 0.1682314 & 0.0073682 & 7 \\
    ePL-KRLS-FSM\_giq (\citealt{marques2022kernel}) & 0.0287900 & 0.1693932 & 0.0074273 & 8 \\
    ePL-KRLS-T1FSM\_mc & 4.0799243 & 2.0165130 & 0.0920681 & 6 \\
    ePL-KRLS-T2FSM\_zl & \textbf{0.0258646} & \textbf{0.1605565} & \textbf{0.0070144} & \textbf{1} \\
    ePL-KRLS-T2FSM\_jg2 & 0.0262004 & 0.1615953 & 0.0070598 & 17 \\
    ePL-KRLS-T2FSM\_zc & 8.8281205 & 2.9662592 & 0.1425426 & 16 \\
    ePL-KRLS-T2FSM\_hc & 4.1844823 & 2.0421885 & 0.0936928 & 16 \\
    ePL-KRLS-T2FSM\_yl & 0.0261339 & 0.1613901 & 0.0070464 & 17 \\
    ePL-KRLS-T2FSM\_ma & \textbf{0.0258646} & \textbf{0.1605565} & \textbf{0.0070144} & \textbf{1} \\
    ePL-KRLS-T2FSM\_hy & \textbf{0.0258646} & \textbf{0.1605565} & \textbf{0.0070144} & \textbf{1} \\
\hline 
\end{tabular}}
\end{center}
\label{TAIEX}
\end{table*}

The results demonstrate that the ePL-KRLS-FSM+ is an efficient model for predicting the main stock index (chaotic time series). Most of the distance measures proposed by the model (ePL-KRLS-FSM\_r2, ePL-KRLS-FSM\_cr and ePL-KRLS-FSM\_giq) obtained similar results with a low number of rules, and the \textit{jaccard\_gt2} and \textit{yang\_lin} T2FSMs also achieve good results.

%% Runtime simulations
\subsection{Runtime simulations}

Table \ref{runtime} shows the runtime of all benchmarks. The values highlighted in blue represent the lowest errors of the model. In general, the Mackey-Glass demonstrated a longer compile time compared to the Stock Market due to the required calculations of the delay, and the outputs being more chaotic and harder to predict.

Long simulation runtimes of the ePL-KRLS-FSM models are expected, as it consists of three steps: participatory learning (PL), kernel recursively least squares (KRLS) method, and the data transformation into FSs. This makes the model more robust than the others, presenting more accurate results.

The models that presented the best runtimes were from the \textit{scikit-learn} library (\citealt{scikit-learn}) (DT, MLP and SVM), which is expected. The ePL-KRLS also presented a very short runtime compared to the other models.

The models that use the \textit{jaccard\_it2} measure (\textit{jaccard\_gt2}, \textit{zhao\_crisp}, \textit{hao\_crisp}) presented execution times almost 10 times longer than the others models. The \textit{yang\_lin} also presented similar runtimes. The ePL-KRLS-T1FSM\_mc model presented the best runtime compared to the FSM models.

The \textit{zl}, \textit{ma} and \textit{hy} T2FSMs presented the lowest errors for the Mackey-Glass with $\vartheta = 500$ and TAIEX, and just one final rule for all simulations. For the ePL-KRLS-FSM+ models, these measures presented the lowest runtime.

Even though type-2 fuzzy sets require more computational time than type-1 (\citealt{zhao2014new}), some models that use this data transformation showed shorter simulation times than the \textit{pappis1} and \textit{zwick} measures. This is due both to the methodology used to calculate the compatibility measure and to the interactions that occur in the other processes of the model.

The ePL-KRLS-FSM\_giq obtained the fifth best runtime for the Mackey-Glass with $\vartheta = 17$, and the best results. The compile time is similar to ePL-KRLS-FSM\_r2 and ePL-KRLS-FSM\_cr models among all simulations.

The \textit{zeng\_li}, \textit{mohamed\_abdaala} and \textit{hung\_yang} measures presented the three best simulation times considering all type-2 metrics, achieving the best results for the Mackey-Glass with $\vartheta = 500$ and TAIEX time series. The \textit{pappis3} also obtained the same values for the errors as these metrics for the main stock index, having the eighth best runtime considering all the twenty models.

Disregarding the simulation time of \textit{scikit-learn} models due to being a very well consolidated and optimized library, the ePL-KRLS-FSM\_r2 model presented the fourth best runtime of all simulations for the Mackey-Glass with $\vartheta = 1000$, where this model presents the best results in terms of errors.

\begin{table*}[htbp]
\caption{Simulations results of runtime time series}
\begin{center}
\resizebox{\linewidth}{!}{%
\begin{tabular}{ c | c c c c }
\hline
    Measure & MG ($\vartheta = 17$) & MG ($\vartheta = 500$) & MG ($\vartheta = 1000$) & TAIEX \\
\hline
    DT (\citealt{breiman1984classification}) & \textbf{2.22 $\pm$ 0.16} & 2.14 $\pm$ 0.23 & \textbf{2.09 $\pm$ 0.33} & 0.05 $\pm$ 0.02 \\
    MLP (\citealt{rumelhart1986learning}) & 2.59 $\pm$ 0.18 & 2.83 $\pm$ 0.14 & 2.69 $\pm$ 0.22 & 0.51 $\pm$ 0.39 \\
    SVM (\citealt{smola2004tutorial}) & 2.31 $\pm$ 0.21 & \textbf{2.02 $\pm$ 0.24} & 2.31 $\pm$ 0.29 & \textbf{0.04 $\pm$ 0.02}\\
    eMG (\citealt{lemos2010multivariable}) & 203.08 $\pm$ 7.46 & 266.68 $\pm$ 10.53 & 357.27 $\pm$ 17.27 & 11.45 $\pm$ 0.38 \\
    ePL-KRLS (\citealt{maciel2017evolving}) & 9.70 $\pm$ 0.35 & 10.69 $\pm$ 0.15 & 10.65 $\pm$ 0.20 & 1.43 $\pm$ 0.67 \\
    ePL-KRLS-DISCO (\citealt{alves2021novel}) & 157.60 $\pm$ 2.47 & 154.67 $\pm$ 3.81 & 155.81 $\pm$ 1.30 & 108.21 $\pm$ 5.82 \\
    ePL-KRLS-FSM\_p1 (\citealt{de2022kernel}) & 588.39 $\pm$ 5.22 & 731,46 $\pm$ 3.31 & 762.42 $\pm$ 7.04 & 265.16 $\pm$ 12.49 \\
    ePL-KRLS-FSM\_p3 (\citealt{de2022kernel}) & 209.21 $\pm$ 0.85 & 302.89 $\pm$ 1.13 & 450.74 $\pm$ 0.66 & \textbf{\textbl{320.70 $\pm$ 8.57}} \\
    ePL-KRLS-FSM\_zw (\citealt{de2022kernel}) & 596.34 $\pm$ 12.34 & 605.98 $\pm$ 43.14 & 781.30 $\pm$ 3.90 & 322.23 $\pm$ 14.22 \\
    ePL-KRLS-FSM\_r2 (\citealt{marques2022kernel}) & 246.11 $\pm$ 9.27 & 258.03 $\pm$ 6.46 & \textbf{\textbl{254.72 $\pm$ 15.87}} & 161.56 $\pm$ 12.31 \\
    ePL-KRLS-FSM\_cr (\citealt{marques2022kernel}) & 248.81 $\pm$ 7.98 & 250.14 $\pm$ 8.53 & 243.07 $\pm$ 12.01 & 166.64 $\pm$ 4.25 \\
    ePL-KRLS-FSM\_giq (\citealt{marques2022kernel}) & \textbf{\textbl{252.05 $\pm$ 5.82}} & 265.62 $\pm$ 13.76 & 260.17 $\pm$ 6.09 & 166.39 $\pm$ 4.90 \\
    ePL-KRLS-T1FSM\_mc & 132.79 $\pm$ 3.72 & 89.74 $\pm$ 2.41 & 91.50 $\pm$ 2.87 & 134.76 $\pm$ 3.26 \\
    ePL-KRLS-T2FSM\_zl & 298.01 $\pm$ 7.96 & \textbf{\textbl{451.22 $\pm$ 9.94}} & 590.67 $\pm$ 8.21 & \textbf{\textbl{333.97 $\pm$ 8.42}} \\
    ePL-KRLS-T2FSM\_jg2 & 3271.21 $\pm$ 131.49 & 6116.68 $\pm$ 213.59 & 6160.20 $\pm$ 256.10 & 1675.94 $\pm$ 56.41\\
    ePL-KRLS-T2FSM\_zc & 3494.76 $\pm$ 151.47 & 5266.89 $\pm$ 178.22 & 5318.85 $\pm$ 182.34 & 1920.28 $\pm$ 79.06 \\
    ePL-KRLS-T2FSM\_hc & 3538.59 $\pm$ 154.09 & 6221.06 $\pm$ 215.71 & 6091.47 $\pm$ 213.94 & 1993.24 $\pm$ 90.38 \\
    ePL-KRLS-T2FSM\_yl & 4040.95 $\pm$ 192.15 & 6444.23 $\pm$ 217.10 & 7009.03 $\pm$ 238.09 & 2423.62 $\pm$ 101.33 \\
    ePL-KRLS-T2FSM\_ma & 458.47 $\pm$ 10.61 & \textbf{\textbl{642.77 $\pm$ 14.02}} & 770.92 $\pm$ 16.31 & \textbf{\textbl{415.14 $\pm$ 9.59}} \\
    ePL-KRLS-T2FSM\_hy & 400.59 $\pm$ 11.65 & \textbf{\textbl{604.25 $\pm$ 13.19}} & 742.93 $\pm$ 15.82 & \textbf{\textbl{384.37 $\pm$ 10.08}} \\
\hline 
\end{tabular}}
\end{center}
\label{runtime}
\end{table*}

%% Overall Discussion
\subsection{Overall Discussion}

The simulations suggest that the ePL-KRLS-FSM+ is a powerful model able to predict chaotic time series, with high accuracy, achieving accurate results and the lowest number of final rules among all models.

Considering all the results, it is possible to notice that some metrics for type-2 fuzzy sets have similarities, among them: the \textit{zeng\_li}, \textit{mohamed\_abdaala}, and \textit{hung\_yang} measures obtained the same error values and number of final rules in all simulations, furthermore, the measures that use the \textit{jaccard\_it2} similarity presented longer runtime compared to others models.

The Tagaki-Sugeno inference method uses polynomial functions in the consequent part, which produces more accurate results with a low number of rules. Polynomial functions can simplify the explanation of complex situations. Consequently, we were able to reduce the number of rules while still obtaining accurate results. The proposed system adapts to the data, indicating that the number of final rules does not necessarily affect the system's explainability; a single final rule can encompass the entire system. This is evident in the ePL-KRLS-T2FSM\_zl, ePL-KRLS-T2FSM\_ma, and ePL-KRLS-T2FSM\_hy models, where having just one final rule yields more accurate results compared to models with more rules.

The ePL-KRLS-FSM+ model is a robust tool for forecasting problems, due to different ways of handling data. Different metrics produce different results depending on the application, It is important to correctly define the parameters for generating FSs in order to optimize the runtime simulations. In general, with less chaotic benchmarks, the FSM models with shorter simulation times (compared with other FSM models) presented more accurate results, and with the increase of chaoticity of data, longer runtimes provide more accurate outputs.

It can also be noted that each metric performs differently in each application, considering the following factors: precision, runtime, and number of final rules. Each measure operates differently, providing unique calculations, rules, and results. This means that it's not just one model that will have the best results for all the experiments.

The opportunity to create different membership functions for type-1 and type-2 fuzzy sets, and compare them across several metrics in the compatibility measure, makes the ePL-KRLS-FSM+ a powerful model for dealing with various applications and benchmarks. Generating fuzzy sets from data, comparing them through measures, and using an eFS to predict the results has demonstrated to be a powerful Machine Learning model, going beyond other eFS and traditional models.

%% Conclusion
\section{Conclusion}\label{Conclusion}

This paper presented an improvement from an existing evolving participatory learning fuzzy model, the ePL-KRLS-FSM, that generates FSs. The new proposal intends to create T2 FSs and compare them through similarity measures. Furthermore, another distance metric for T1 fuzzy sets is added, and the introduction of entropy and inclusion metrics is discussed. There are many forecasting models for predicting chaotic data, so ePL-KRLS-FSM+ stands out by being capable of using different methods for calculating the compatibility measure, an important part of the ePL models. The Mackey-Glass chaotic time series with different chaos levels, and the TAIEX main stock index are used to evaluate the efficiency of the proposed model. Most of the new measures adopted showed satisfactory results in the applications, obtaining better outputs than the type-1 models for $\vartheta = 500$, and matching the best model for TAIEX. The Results indicate that the ePL-KRLS-FSM+ is a robust model able to adapt its functionality and structure with the exchanging of data, precisely predicting data and with a competitive computational cost. 

For future work, we propose to apply a hyperparameter optimization due to the increase of parameters (membership functions for type-1 and type-2 fuzzy sets, and 
countless metrics for the compatibility measure) to control the learning process. Further, due to the increased modeling capabilities of type-2 fuzzy logic for handling uncertainty, we suggest different and extended designs for modeling T2 fuzzy sets and the possibility of changing uncertainty levels.

%\begin{acknowledgements}
%\end{acknowledgements}

\bibliographystyle{spbasic}      % basic style, author-year citations
\bibliography{svbib}   % name your BibTeX data base

\end{sloppypar}
\end{document}